\documentclass{article}
\usepackage{microtype}
\usepackage{graphicx}
\usepackage{subfigure}
\usepackage{amsfonts}
\usepackage{commath}
\usepackage{amsmath}
\usepackage{multirow}
\usepackage{dirtytalk}
\usepackage{mathtools}
\usepackage[shortlabels]{enumitem}
\newcommand\independent{\protect\mathpalette{\protect\independenT}{\perp}}
\def\independenT#1#2{\mathrel{\rlap{$#1#2$}\mkern2mu{#1#2}}}
\usepackage{amsthm}

\DeclareMathOperator*{\argmin}{arg\,min}
\newtheorem{theorem}{Theorem}[section]
\newtheorem{lemma}[theorem]{Lemma}

\newtheorem{innercustomproof}{}
\newenvironment{customproof}[1]
{\renewcommand\theinnercustomproof{#1}\innercustomproof}
{\endinnercustomproof}
\theoremstyle{definition}
\newtheorem{definition}[theorem]{Definition}

\usepackage{tikz-cd}
\usepackage{booktabs} 
\usepackage{bm}
\allowdisplaybreaks

\usepackage{hyperref}

\usepackage[accepted]{icml2021}

\icmltitlerunning{CoDiTE with CMEs and U-Statistic Regression}

\begin{document}
	
	\twocolumn[
	\icmltitle{Conditional Distributional Treatment Effect with Kernel Conditional Mean Embeddings and U-Statistic Regression}
	
	
	
	\icmlsetsymbol{equal}{*}
	
	\begin{icmlauthorlist}
		\icmlauthor{Junhyung Park}{mpi}
		\icmlauthor{Uri Shalit}{technion}
		\icmlauthor{Bernhard Sch\"olkopf}{mpi}
		\icmlauthor{Krikamol Muandet}{mpi}
	\end{icmlauthorlist}
	
	\icmlaffiliation{mpi}{Max Planck Institute for Intelligent Systems, T\"ubingen, Germany}
	\icmlaffiliation{technion}{Technion, Israel Institute of Technology}
	
	\icmlcorrespondingauthor{Junhyung Park}{junhyung.park@tuebingen.mpg.de}
	
	\icmlkeywords{kernels, RKHS, kernel mean embedding, conditional mean embedding, individualised treatment effect, causal inference, conditional two-sample testing, conditional average treatment effect, conditional distributional treatment effect, U-statistics, conditional U-statistics, U-statistic Regression}
	
	\vskip 0.3in
	]

	
	
	\printAffiliationsAndNotice{}  
	
	\begin{abstract}
		We propose to analyse the conditional distributional treatment effect (CoDiTE), which, in contrast to the more common conditional average treatment effect (CATE), is designed to encode a treatment's distributional aspects beyond the mean. We first introduce a formal definition of the CoDiTE associated with a distance function between probability measures. Then we discuss the CoDiTE associated with the maximum mean discrepancy via kernel conditional mean embeddings, which, coupled with a hypothesis test, tells us whether there is any conditional distributional effect of the treatment. Finally, we investigate what kind of conditional distributional effect the treatment has, both in an exploratory manner via the conditional witness function, and in a quantitative manner via U-statistic regression, generalising the CATE to higher-order moments. Experiments on synthetic, semi-synthetic and real datasets demonstrate the merits of our approach. 
	\end{abstract}
	
	\section{Introduction}
	Analysing the effect of a treatment (medical drug, economic programme, etc.) has long been a problem of great importance, and has attracted researchers from diverse domains, including econometrics \cite{imbens2009recent}, political sciences \cite{kunzel2019metalearners}, healthcare \cite{foster2011subgroup} and social sciences \cite{imbens2015causal}. The field has naturally received much attention of statisticians over the years \cite{rosenbaum2002observational,rubin2005causal,imbens2015causal}, and in the past few years, the machine learning community has started applying its own armoury to this problem -- see Section \ref{SSrelatedworks} for a succinct review. 
	
	Traditional methods for treatment effect evaluation focus on the analysis of the average treatment effect (ATE), such as an increase or decrease in average income, inequality or poverty, aggregated over the population. However, the ATE is not informative about the individual responses to the intervention and how the treatment impact varies across individuals (known as \textit{treatment effect heterogeneity}). The study of conditional average treatment effect (CATE)\footnote{See Section \ref{SSproblem} for the definitions of the ATE and CATE.} has been proposed to analyse such heterogeneity in the mean treatment effect. 
	Although sufficient in many cases, the CATE is still an average. As such, it fails to capture information about distributional aspects of the treatment beyond the mean.
	A significant amount of interest exists for developing methods that can analyse distributional treatment effects conditioned on the covariates \cite{chang2015nonparametric,bitler2017can,shen2019estimation,chernozhukov2020network,hohberg2020treatment,briseno2020flexible}. 
	
	Our contributions are as follows. Firstly, we formally define the \textit{conditional distributional treatment effect} (CoDiTE) associated with a chosen distance function between distributions.
	Then we use kernel conditional mean embeddings \cite{song2013kernel,park2020measure} to analyse the CoDiTE associated with the \textit{maximum mean discrepancy} \cite{gretton2012kernel}. Coupled with a statistical hypothesis test, this can determine \textit{whether} there exists any effect of the treatment, conditioned on a set of covariates. Finally, we use \textit{conditional witness functions} and \textit{U-statistic regression} to investigate \textit{what kind} of effect the treatment has. 
	
	\begin{figure*}[t]
		\centerline{\includegraphics[scale=0.43]{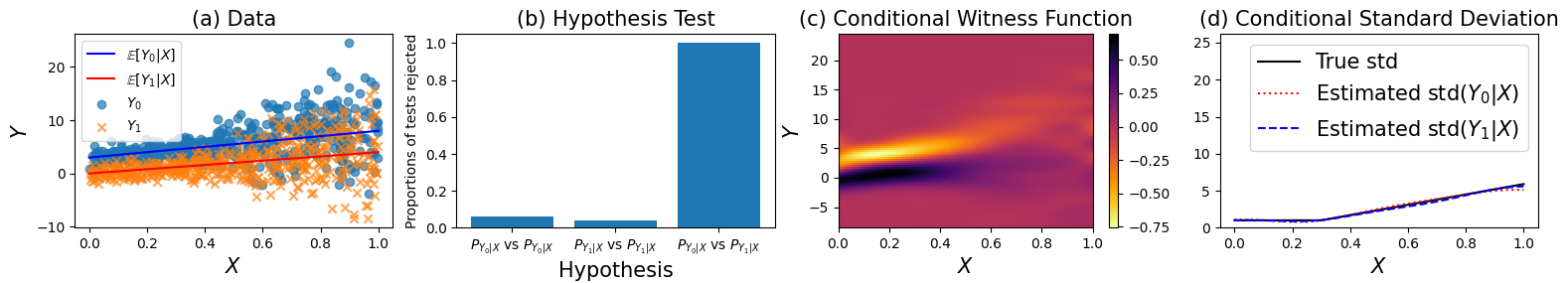}}
		\caption{\textbf{Toy illustration of higher-order heterogeneity that cannot be captured by CATE.} \textbf{(a) Data.} \(X\sim\text{Uniform}[0,1]\), \(Y_0=3+5X+\mathbf{1}_{X<0.3}N+7\mathbf{1}_{X\geq0.3}(1+(X-0.3))N\) and \(Y_1=4X+\mathbf{1}_{X<0.3}N+7\mathbf{1}_{X\geq0.3}(1+(X-0.3))N\), where \(N\sim\mathcal{N}(0,1)\); in particular, the CATE is increasing with \(X\). \textbf{(b) Hypothesis test} (Section \ref{SStest}) Each of the hypotheses \(P_{Y_0|X}\equiv P_{Y_0|X}\), \(P_{Y_1|X}\equiv P_{Y_1|X}\) and \(P_{Y_0|X}\equiv P_{Y_1|X}\) are tested 100 times. The last (false) hypothesis is rejected in most tests, while the first two (true) hypotheses are not rejected in most tests, meaning that both type I and type II errors are low. \textbf{(c) Conditional witness function} (Section \ref{SSwitness}). The conditional witness function is close to zero for all \(Y\) at \(X\geq0.5\), demonstrating that \(P_{Y_0|X}\) and \(P_{Y_1|X}\) are similar in this region of \(\mathcal{X}\). For \(X<0.4\), the witness function is positive in regions where the density of \(Y_1\) is higher than that of \(Y_0\), and negative in regions where the density of \(Y_0\) is higher than that of \(Y_1\). \textbf{(d) U-statistic regression} (Section \ref{SScoditemoments}). True conditional standard deviation (in black) is estimated (in red and blue for control and treatment groups respectively) as a function of \(X\) via U-statistic regression (since variance is a U-statistic) and the square-root operation. We see that the standard deviation increases linearly for \(X\geq0.3\).}
		\label{Fmotivation}
	\end{figure*}
	
	\subsection{Problem Set-Up: Potential Outcomes Framework}\label{SSproblem}
	
	Throughout this paper, we take \((\Omega,\mathcal{F},P)\) as the underlying probability space, \(\mathcal{X}\) as the input space and \(\mathcal{Y}\subseteq\mathbb{R}\) as the output space. Let \(Z:\Omega\rightarrow\{0,1\}\), \(X:\Omega\rightarrow\mathcal{X}\) and \(Y_0,Y_1,Y:\Omega\rightarrow\mathcal{Y}\) be random variables representing, respectively, the treatment assignment, covariates, the potential outcomes under control and treatment, and the observed outcome,
	i.e. \(Y=Y_0(1-Z)+Y_1Z\). For example, \(Z\) may indicate whether a subject is administered a medical treatment (\(Z=1\)) or not (\(Z=0\)). The potential outcomes \(Y_1,Y_0\) respectively correspond to subject's responses had they received treatment or not. The covariates \(X\) correspond to subject's characteristics such as age, gender, race that could influence both the potential outcomes and the choice of treatment. We denote the distributions of random variables by subscripting \(P\), e.g. \(P_X\) for the distribution of \(X\). Throughout, we impose the mild condition that conditional distribution \(P(\cdot\mid X)\) admits a \textit{regular version} \citep[p.150, Definition 2.4, Proposition 2.5]{cinlar2011probability}. 
	
	Each unit \(i=1,...,n\) is associated with an independent copy \((X_i,Z_i,Y_{0i},Y_{1i})\) of \((X,Z,Y_0,Y_1)\). However, for each \(i=1,...,n\), we observe either \(Y_{0i}\) or \(Y_{1i}\); this missing value problem is known as the \textit{fundamental problem of causal inference} \citep{holland1986statistics}, preventing us from directly computing the difference in the outcomes under treatment and control for each unit. As a result, we only have access to samples \(\{(x_i,z_i,y_i)\}^n_{i=1}\) of \((X,Z,Y)\).  We write \(n_0=\sum^n_{i=1}\mathbf{1}_{z_i=0}\) and \(n_1=\sum^n_{i=1}\mathbf{1}_{z_i=1}\) for the control and treatment sample sizes, and denote the control and treatment samples by \(\{(x^0_i,y^0_i)\}^{n_0}_{i=1}\) and \(\{(x^1_i,y^1_i)\}^{n_1}_{i=1}\).
	
	We assume \textit{strong ignorability} \cite{rosenbaum1983central}:
	\begin{description}
		\itemsep0em 
		\item[unconfoundedness] \(Z\independent(Y_0,Y_1)\mid X\); \;\; and
		\item[overlap] \(0<e(X)=P(Z=1\mid X)=\mathbb{E}[Z\mid X]<1\). 
	\end{description}
	Causal treatment effects are then identifiable from observational data, since \(P_{Y_0|X}=P_{Y_0|X,Z=0}=P_{Y|X,Z=0}\), and similarly for \(P_{Y_1|X}\). The quantity \(e(X)\) is the \textit{propensity score}. 
	In a \textit{randomised experiment}, \(e(X)\) is known and controlled \citep[p.40, Definition 3.10]{imbens2015causal}. 
	
	The usual objects of interest in the treatment effect literature are the \textit{average treatment effect} (ATE), \(\mathbb{E}\left[Y_1-Y_0\right]\),
	and the \textit{conditional average treatment effect} (CATE), \(T(x)=\mathbb{E}\left[Y_1-Y_0\mid X=x\right]\). In this paper, we propose to extend the analysis to compare other aspects of the conditional distributions, \(P_{Y_0|X}\) and \(P_{Y_1|X}\). One compelling reason to do this is that estimating CATE is inherently a problem of \textit{comparing two means}, and as such, is only meaningful if the corresponding variances are given. Consider the toy example in Figure \ref{Fmotivation}. The CATE is constructed to be increasing with \(X\), but taking into account the variance, the treatment effect is clearly more pronounced for small values of \(X\). For example, the probability of $Y_1$ being greater than $Y_0$ is much higher for smaller values of \(X\). 
	
	Beyond the mean and variance, researchers may also be interested in other higher-moment treatment effect heterogeneity, such as Gini's mean difference or skewness, or indeed how the entire conditional densities of the control and treatment groups differ given the covariates, in an exploratory fashion. Panels (b), (c) and (d) in Figure \ref{Fmotivation} demonstrate each of the steps we propose in this paper applied to this toy dataset: hypothesis testing of equality of conditional distributions, the conditional witness function and U-statistic regression (variance, in this instance), respectively. 
	
	\subsection{Related Work \& Summary of Contributions}\label{SSrelatedworks}
	In the past few years the machine learning community has focused much effort on models for estimating the CATE function. Some approaches include Gaussian processes \cite{alaa2017bayesian,alaa2018bayesian},	Bayesian regression trees \cite{hill2011bayesian,hahn2020bayesian}, random forests \cite{wager2018estimation}, neural networks \cite{johansson2016learning,shalit2017estimating,louizos2017causal,atan2018deep,shi2019adapting}, GANs \cite{yoon2018ganite}, boosting and adaptive regression splines \cite{powers2018some} and kernel mean embeddings \cite{singh2020kernel}.
	
	Distributional extensions of the ATE have been considered by many authors. \citet{abadie2002bootstrap} tested the hypotheses of equality and stochastic dominance of the marginal outcome distributions \(P_{Y_0}\)	and \(P_{Y_1}\), whereas \citet{kim2018causal,muandet2020counterfactual} focus on estimating \(P_{Y_0}\)	and \(P_{Y_1}\), or some distance between them. These works do not consider treatment effect heterogeneity. \citet[Appendix C]{singh2020kernel} consider CATE as well as distributional treatment effect,
	and while it seems that the ideas can straightforwardly be extended to conditional distributional treatment effect, it is
	not explicitly considered in the paper.
	
	The CoDiTE incorporates both distributional considerations of treatment effects \textit{and} treatment effect heterogeneity. Interest has been growing, especially in the econometrics literature, for such analyses -- indeed, \citet{bitler2017can} provided concrete evidence that in some settings, the CATE does not suffice. Existing works that analyse the CoDiTE can be split into three categories, depending on how distributions are characterised: (i) quantiles, (ii) cumulative distributional functions, and (iii) specific distributional parameters, such as the mean, variance, skewness, etc. In category (i), quantile regression is a powerful tool \cite{koenker2005quantile}; however, in order to get a distributional picture via quantiles, one needs to estimate a large number of quantiles, and issues of crossing quantiles arise, whereby estimated quantiles are non-monotone. In category (ii), \citet{chernozhukov2013inference,chernozhukov2020network} propose splitting \(\mathcal{Y}\) into a grid and regressing for the cumulative distribution function at each point in the grid, but this also brings issues of non-monotonicity of the cumulative distribution function, similar to crossing quantiles. \citet{shen2019estimation} estimates the cumulative distribution functions \(P(Y_0<y^*)\) and \(P(Y_1<y^*)\) for each \(y^*\in\mathcal{Y}\) given each value of \(X=x\) by essentially applying the Nadaraya-Watson conditional U-statistic of \citet{stute1991conditional} to the U-kernel \(h(y)=\mathbf{1}(y\leq y^*)\). In category (iii), generalised additive models for location, scale and shape (GAMLSS) \cite{stasinopoulos2017flexible} have been applied for CoDiTE analysis \cite{hohberg2020treatment,briseno2020flexible}, but being a parametric model, despite its flexibility, the researcher has to choose a model beforehand to proceed, and issues of model misspecification are unavoidable. 
	
	Interest has also always existed for hypothesis tests in the context of treatment effect analysis, especially in econometrics \citep[Sections 3.3 and 5.12]{imbens2009recent}. \citet{abadie2002bootstrap} tested the equality between the marginal distributions of \(Y_0\) and \(Y_1\), while \citet{crump2008nonparametric} tested for the equality of \(\mathbb{E}[Y_1|X]\) and \(\mathbb{E}[Y_0|X]\). \citet{lee2009nonparametric,lee2009non,chang2015nonparametric,shen2019estimation} were interested, among others, in the hypothesis of the equality of \(P_{Y_1|X}\) and \(P_{Y_0|X}\), which we consider in Section \ref{SStest}. 
	
	\paragraph{Summary of Contributions}
	We characterise distributions in two ways -- first as elements in a reproducing kernel Hilbert space via kernel conditional mean embeddings, which, to the best of our knowledge, is a novel attempt in the treatment effect literature, and secondly via specific distributional parameters, as in category (iii). The former characterisation gives us a novel way of testing for the equality of conditional distributions, as well as an exploratory tool for density comparison between the groups via conditional witness functions. For the latter characterisation, we provide, to the best of our knowledge, a novel U-statistic regression technique by generalising kernel ridge regression, which, in contrast to GAMLSS, is fully nonparametric. Neither characterisation requires the estimation of a large number of quantities, unlike characterisations via quantiles or cumulative distribution functions. 
	
	\section{Preliminaries}\label{Spreliminaries}
	In this section, we briefly review reproducing kernel Hilbert space embeddings and U-statistics. A more complete introduction can be found in Appendix \ref{Smoredetails}. 
	
	\subsection{Reproducing Kernel Hilbert Space Embeddings}\label{SSrkhs}
	Let \(l:\mathcal{Y}\times\mathcal{Y}\rightarrow\mathbb{R}\) be a (scalar) positive definite kernel on \(\mathcal{Y}\) with \textit{reproducing kernel Hilbert space} (RKHS) \(\mathcal{H}\) \citep[p.7, Definition 1]{berlinet2004reproducing}. Given a random variable \(Y\) on \(\mathcal{Y}\) satisfying \(\mathbb{E}[\sqrt{l(Y,Y)}]<\infty\), the \textit{kernel mean embedding} of \(Y\) is defined as \(\mu_Y(\cdot)=\mathbb{E}[l(Y,\cdot)]\) \citep[Eqn. (2a)]{smola2007hilbert}. Given two random variables \(Y\) and \(Y'\), the \textit{maximum mean discrepancy} (MMD) between them is defined as \(\lVert\mu_Y-\mu_{Y'}\rVert_\mathcal{H}\) \citep[Lemma 4]{gretton2012kernel}, where \(\mu_Y-\mu_{Y'}\) is the (unnormalised) \textit{witness function} (\citealp[Section 2.3]{gretton2012kernel}; \citealp[Eqn. (3.2)]{lloyd2015statistical}). If the embedding is injective from the space of probability measures on \(\mathcal{Y}\) to \(\mathcal{H}\), then we say that \(l\) is \textit{characteristic} \citep[Section 2.2]{fukumizu2008kernel}, in which case the MMD is a proper metric. Given another random variable \(X\) on \(\mathcal{X}\), the \textit{conditional mean embedding} (CME) of \(Y\) given \(X\) is defined as \(\mu_{Y|X}=\mathbb{E}[l(Y,\cdot)\mid X]\) \citep[Definition 3.1]{park2020measure}\footnote{We use the conditional expectation interpretation of the CME. An interpretation of the CME as an operator from an RKHS on \(\mathcal{X}\) to \(\mathcal{H}\) also exists \cite{song2009hilbert,song2013kernel,fukumizu2013kernel}.}. 
	
	Denote by \(L^2(\mathcal{X},P_X;\mathcal{H})\) the Hilbert space of (equivalence classes of) measurable functions \(F:\mathcal{X}\rightarrow\mathcal{H}\) such that \(\lVert F(\cdot)\rVert^2_{\mathcal{H}}\) is \(P_X\)-integrable, with inner product 
	\(\langle F_1,F_2\rangle_2=\int_\mathcal{X}\langle F_1(x),F_2(x)\rangle_{\mathcal{H}}dP_X(x)\). Given an \textit{operator-valued kernel} \(\Gamma:\mathcal{X}\times\mathcal{X}\rightarrow\mathcal{L}(\mathcal{H})\), where \(\mathcal{L}(\mathcal{H})\) is the Banach space of bounded linear operators \(\mathcal{H}\rightarrow\mathcal{H}\), there exists an associated \textit{vector-valued RKHS} of functions \(\mathcal{X}\rightarrow\mathcal{H}\) \citep[Definition 2.1, Definition 2.2, Proposition 2.3]{carmeli2006vector}. 
	
	\subsection{U-Statistics}\label{SSustatistics}
	Let \(Y_1,...,Y_r\) be independent copies of \(Y\), and let \(h:\mathcal{Y}^r\rightarrow\mathbb{R}\) be a symmetric function, i.e. for any permutation \(\pi\) of \((1,...,r)\), \(h(y_1,...,y_r)=h(y_{\pi(1)},...,y_{\pi(r)})\), such that \(h(Y_1,...,Y_r)\) is integrable. Given i.i.d. copies \(\{Y_i\}_{i=1}^n\) of \(Y\), the \textit{U-statistic} \citep[p. 172]{hoeffding1948class,serfling1980approximation} for an unbiased estimation of \(\theta(P_Y)=\mathbb{E}[h(Y_1,...,Y_r)]\) is \(\hat{\theta}(Y_1,...,Y_n)=\frac{1}{\binom{n}{r}}\sum h\left(Y_{i_1},...,Y_{i_r}\right)\) where \(\binom{n}{r}\) is the binomial coefficient and the summation is over the \(\binom{n}{r}\) combinations of \(r\) distinct elements \(\{i_1,...,i_r\}\) from \(\{1,...,n\}\). 
	
	This has been extended to the conditional case \cite{stute1991conditional}. Given another random variable \(X\) on \(\mathcal{X}\) and independent copies \(X_1,...,X_r\) of it, we can consider the estimation of \(\theta(P_{Y|X})=\mathbb{E}[h(Y_1,...,Y_r)|X_1,...,X_r]\). \citet{stute1991conditional,derumigny2019estimation} extend the Nadaraya-Watson regressor  \cite{nadaraya1964estimating,watson1964smooth} to estimate \(\theta(P_{Y|X})\). 
	
	\section{Conditional Distributional Treatment Effect}\label{Scodite}
	In this section, we generalise the notion of CATE to account for distributional differences between treatment and control groups, rather than just the mean difference. 
	
	\begin{definition}\label{Dcodite}
		Let \(D\) be some distance function between probability measures. We define the \textit{conditional distributional treatment effect} (CoDiTE) associated with \(D\) as
		\[U_D(x)=D(P_{Y_0|X=x},P_{Y_1|X=x}).\]
	\end{definition}
	Here, the choice of \(D\) depends on what characterisation of distributions is used (c.f. Section \ref{SSrelatedworks}). For example, if \(D(P_{Y_0|X=x},P_{Y_1|X=x})=\mathbb{E}[Y_1\mid X=x]-\mathbb{E}[Y_0\mid X=x]\), we recover the CATE, i.e. \(U_D(x)=T(x)\), thereby showing that the CoDiTE is a strict generalisation of the CATE. Different choices of \(D\) will require different estimators. 
	
	The usual performance metric of a CATE estimator \(\hat{T}\) is the \textit{precision of estimating heterogeneous effects} (PEHE) (first proposed in sample form by \citet[Section 4.3]{hill2011bayesian}; we report the population-level definition, found in, for example, \citet[Eqn. (5)]{alaa2019validating}:
	\[\lVert\hat{T}-T\rVert^2_2=\mathbb{E}[\lvert\hat{T}(X)-T(X)\rvert^2].\]
	We propose a performance metric of an estimator of the CoDiTE in an exactly analogous manner. 
	\begin{definition}\label{Dpehde}
		Given a distance function \(D\), for an estimator \(\hat{U}_D\) of \(U_D\), we define the \textit{precision of estimating heterogeneous distributional effects} (PEHDE) as
		\[\psi_D(\hat{U}_D)=\lVert\hat{U}_D-U_D\rVert_2^2=\mathbb{E}[\lvert\hat{U}_D(X)-U_D(X)\rvert^2].\]
	\end{definition}
	Again, if \(D\) measures the difference in expectations, then the associated PEHDE \(\psi_D\) reduces to the usual PEHE. 
	
	Henceforth, we explore different choices of the distance function \(D\), as well as methods of estimating the corresponding CoDiTE \(U_D\), to answer the following questions: 
	\begin{description}
		\item[Q1] Are \(P_{Y_0|X}\) and \(P_{Y_1|X}\) different? In other words, is there any distributional effect of the treatment? (Section \ref{Scoditemmd})
		\item[Q2] If so, how does the distribution of the treatment group differ from that of the control group? (Section \ref{Sunderstanding})
	\end{description}
	
	\section{CoDiTE associated with MMD via CMEs}\label{Scoditemmd}
	In this section, we answer Q1, i.e. we investigate whether the treatment has any effect at all. To this end we choose \(D\) to be the MMD with the associated kernel \(l\) being characteristic. Then writing \(\mu_{Y_0|X}\) and \(\mu_{Y_1|X}\) for the CMEs of \(Y_0\) and \(Y_1\) given \(X\) respectively (c.f. Section \ref{SSrkhs}), we have
	\begin{equation}\label{EcoditeF}
		\begin{split}
			U_\textnormal{MMD}(x)&=\textnormal{MMD}(P_{Y_0|X=x},P_{Y_1|X=x})\\
			&=\lVert\mu_{Y_1|X=x}-\mu_{Y_0|X=x}\rVert_{\mathcal{H}}.
		\end{split}
	\end{equation}
	Since \(l\) is characteristic, \(P_{Y_0|X=x}\) and \(P_{Y_1|X=x}\) are equal if and only if \(\textnormal{MMD}(P_{Y_0|X=x},P_{Y_1|X=x})=0\). What makes the MMD a particularly convenient choice is that for each \(x\in\mathcal{X}\), \(P_{Y_0|X=x}\) and \(P_{Y_1|X=x}\) are represented by individual elements \(\mu_{Y_0|X=x}\) and \(\mu_{Y_1|X=x}\) in the RKHS \(\mathcal{H}\), which means that we can estimate the associated CoDiTE simply by performing regression with \(\mathcal{X}\) as the input space and \(\mathcal{H}\) as the output space, as will be shown in the next section. 
	
	\subsection{Estimation and Consistency}\label{SSestimationmmd}
	We now discuss how to obtain empirical estimates of \(U_\textnormal{MMD}(x)\). Recall that, by the unconfoundedness assumption, we can estimate \(\mu_{Y_0|X}\) and \(\mu_{Y_1|X}\) separately from control and treatment samples respectively. We perform operator-valued kernel regression \cite{micchelli2005learning,kadri2016operator} in separate vector-valued RKHSs \(\mathcal{G}_0\) and \(\mathcal{G}_1\), endowed with kernels \(\Gamma_0(\cdot,\cdot)=k_0(\cdot,\cdot)\text{Id}\) and \(\Gamma_1(\cdot,\cdot)=k_1(\cdot,\cdot)\text{Id}\), where \(k_0,k_1:\mathcal{X}\times\mathcal{X}\rightarrow\mathbb{R}\) are scalar-valued kernel and \(\text{Id}:\mathcal{H}\rightarrow\mathcal{H}\) is the identity operator. Following \citet[Eqn. (4)]{park2020measure}, the empirical estimates \(\hat{\mu}_{Y_0|X}\) and \(\hat{\mu}_{Y_1|X}\) of \(\mu_{Y_0|X}\) and \(\mu_{Y_1|X}\) are constructed, for each \(x\in\mathcal{X}\), as
	\begin{equation}\label{EempiricalF}
		\begin{split}
			\hat{\mu}_{Y_0|X=x}&=\bm{k}_0^T(x)\mathbf{W}_0\bm{l}_0\in\mathcal{G}_0\\
			\text{and}\quad\hat{\mu}_{Y_1|X=x}&=\bm{k}_1^T(x)\mathbf{W}_1\bm{l}_1\in\mathcal{G}_1,\quad\text{where}
		\end{split}
	\end{equation} 
	\(\mathbf{W}_0=(\mathbf{K}_0+n_0\lambda^0_{n_0}\mathbf{I}_{n_0})^{-1}\), \(\mathbf{W}_1=(\mathbf{K}_1+n_1\lambda^1_{n_1}\mathbf{I}_{n_1})^{-1}\), \([\mathbf{K}_0]_{1\leq i,j\leq n_0}=k_0(x^0_i,x^0_j)\), \([\mathbf{K}_1]_{1\leq i,j\leq n_1}=k_1(x^1_i,x^1_j)\), \(\lambda^0_{n_0},\lambda^1_{n_1}>0\) are regularisation parameters, \(\mathbf{I}_{n_0}\) and \(\mathbf{I}_{n_1}\) are identity matrices, \(\bm{k}_0(x)=(k_0(x^0_1,x),...,k_0(x^0_{n_0},x))^T\), \(\bm{k}_1(x)=(k_1(x^1_1,x),...,k_1(x^1_{n_1},x))^T\), \(\bm{l}_0=(l(y^0_1,\cdot),...,l(y^0_{n_0},\cdot))^T\) and \(\bm{l}_1=(l(y^1_1,\cdot),...,l(y^1_{n_1},\cdot))^T\).
	
	By plugging in the estimates (\ref{EempiricalF}) in the expression (\ref{EcoditeF}) for \(U_\textnormal{MMD}\), we can construct \(\hat{U}_\textnormal{MMD}\) as
	\[\hat{U}_\textnormal{MMD}(x)=\lVert\hat{\mu}_{Y_1|X=x}-\hat{\mu}_{Y_0|X=x}\rVert_\mathcal{H}.\]
	The next lemma establishes a closed-form expression for \(\hat{U}_\textnormal{MMD}\) based on the control and treatment samples. 
	\begin{lemma}\label{LhatU}
		For each \(x\in\mathcal{X}\), we have
		\begin{alignat*}{2}
			\hat{U}_\textnormal{MMD}^2(x)&=\bm{k}_0^T(x)\mathbf{W}_0\mathbf{L}_0\mathbf{W}_0^T\bm{k}_0(x)\\
			&\quad-2\bm{k}^T_0(x)\mathbf{W}_0\mathbf{L}\mathbf{W}^T_1\bm{k}_1(x)\\
			&\qquad+\bm{k}_1^T(x)\mathbf{W}_1\mathbf{L}_1\mathbf{W}_1^T\bm{k}_1(x),\quad\text{where}
		\end{alignat*}
		\([\mathbf{L}_0]_{1\leq i,j\leq n_0}=l(y^0_i,y^0_j)\), \([\mathbf{L}]_{1\leq i\leq n_0,1\leq j\leq n_1}=l(y^0_i,y^1_j)\) and \([\mathbf{L}_1]_{1\leq i,j\leq n_1}=l(y^1_i,y^1_j)\).
	\end{lemma}
	The proof of this, and all other results, are deferred to Appendix \ref{Sproofs}. The next theorem shows that, using \textit{universal kernels} \(\Gamma_0,\Gamma_1\) \citep[Definition 4.1]{carmeli2010vector}, \(\hat{U}_\textnormal{MMD}\) is universally consistent with respect to the PEHDE.
	\begin{theorem}[Universal consistency]\label{Tconsistencymmd}
		Suppose that \(k_0,k_1\) and \(l\) are bounded, that \(\Gamma_0\) and \(\Gamma_1\) are universal, and that \(\lambda^0_{n_0}\) and \(\lambda^1_{n_1}\) decay at slower rates than \(\mathcal{O}(n_0^{-1/2})\) and \(\mathcal{O}(n_1^{-1/2})\) respectively. Then as  \(n_0,n_1\rightarrow\infty\), 
		\[\psi_\textnormal{MMD}(\hat{U}_\textnormal{MMD})=\mathbb{E}[(\hat{U}_\textnormal{MMD}(X)-U_\textnormal{MMD}(X))^2]\stackrel{p}{\rightarrow}0.\]
	\end{theorem}
	
	\subsection{Statistical Hypothesis Testing}\label{SStest}
	We are interested in whether or not the two conditional distributions \(P_{Y_0|X}\) and \(P_{Y_1|X}\), corresponding to control and treatment, are equal. The hypotheses are then
	\begin{description}
		\itemsep0em
		\item[\(H_0\):] \(P_{Y_0|X=x}(\cdot)=P_{Y_1|X=x}(\cdot)\) \(P_X\)-almost everywhere. 
		\item[\(H_1\):] There exists \(A\subseteq\mathcal{X}\) with positive measure such that \(P_{Y_0|X=x}(\cdot)\neq P_{Y_1|X=x}(\cdot)\) for all \(x\in A\). 
	\end{description}
	The null hypothesis \(H_0\) means that the treatment has no effect for any of the covariates, whereas the alternative hypothesis \(H_1\) means that the treatment has an effect on \textit{some} of the covariates, where the effect is distributional. For notational simplicity, we write \(P_{Y_0|X}\equiv P_{Y_1|X}\) if \(H_0\) holds.
	
	\begin{algorithm}[tb]
		\caption{Kernel conditional discrepancy (KCD) test of conditional distributional treatment effect}
		\label{Atest}
		\begin{algorithmic}
			\STATE {\bfseries Input:} data \(\{(x_i,z_i,y_i)\}^n_{i=1}\), significant level \(\alpha\), kernels \(k_0,k_1,l\), regularisation parameters \(\lambda^0_{n_0},\lambda^1_{n_1}\), no. of permutations \(m\). 
			\STATE Calculate \(\hat{t}\) using Lemma \ref{Lhatt} based on the input data. 
			\STATE KLR of \(\{z_i\}^n_{i=1}\) against \(\{x_i\}^n_{i=1}\) to obtain \(\hat{e}(x_i)\).
			\FOR{\(k=1\) {\bfseries to} \(m\)}
			\STATE For each \(i=1,...,n\), sample \(\tilde{z}_i\sim\text{Bernoulli}(\hat{e}(x_i))\).
			\STATE Calculate \(\hat{t}_k\) from the new dataset \(\{x_i,\tilde{z}_i,y_i\}_{i=1}^n\).
			\ENDFOR
			\STATE Calculate the \(p\)-value as \(p=\frac{1+\sum^m_{l=1}\mathbf{1}\left\{\hat{t}_l>\hat{t}\right\}}{1+m}\). 
			\IF{\(p<\alpha\)}
			\STATE Reject \(H_0\).
			\ENDIF
		\end{algorithmic}
	\end{algorithm}
	
	We use the following criterion for \(P_{Y_0|X}\equiv P_{Y_1|X}\), which we call the \textit{kernel conditional discrepancy} (KCD):
	\[t=\mathbb{E}[\lVert \mu_{Y_1|X}-\mu_{Y_0|X}\rVert^2_\mathcal{H}].\]
	The following lemma tells us that \(t\) can indeed be used as a criterion of \(P_{Y_0|X}\equiv P_{Y_1|X}\). 
	\begin{lemma}\label{Lequalitycriterion}
		If \(l\) is a characteristic kernel, \(P_{Y_0|X}\equiv P_{Y_1|X}\) if and only if \(t=0\). 
	\end{lemma}
	Next, we define a plug-in estimate \(\hat{t}\) of \(t\), which we will use as the test statistic of our hypothesis test: 
	\[\hat{t}=\frac{1}{n}\sum^n_{i=1}\left\Vert\hat{\mu}_{Y_1|X=x_i}-\hat{\mu}_{Y_0|X=x_i}\right\rVert_\mathcal{H}^2.\]
	Then we have a closed-form expression for \(\hat{t}\) as follows.
	\begin{lemma}\label{Lhatt}
		We have
		\begin{alignat*}{2}
			\hat{t}&=\frac{1}{n}\textnormal{Tr}\left(\tilde{\mathbf{K}}_0\mathbf{W}_0\mathbf{L}_0\mathbf{W}_0^T\tilde{\mathbf{K}}^T_0\right)\\
			&\quad-\frac{2}{n}\textnormal{Tr}\left(\tilde{\mathbf{K}}_0\mathbf{W}_0\mathbf{L}\mathbf{W}^T_1\tilde{\mathbf{K}}^T_1\right)\\
			&\qquad+\frac{1}{n}\textnormal{Tr}\left(\tilde{\mathbf{K}}_1\mathbf{W}_1\mathbf{L}_1\mathbf{W}_1^T\tilde{\mathbf{K}}^T_1\right),
		\end{alignat*}
		where \(\mathbf{L}_0,\mathbf{L}_1\) and \(\mathbf{L}\) are as defined in Lemma \ref{LhatU} and \([\tilde{\mathbf{K}}_0]_{1\leq i\leq n,1\leq j\leq n_0}=k_0(x_i,x^0_j)\) and \([\tilde{\mathbf{K}}_1]_{1\leq i\leq n,1\leq j\leq n_1}=k_1(x_i,x^1_j)\).
	\end{lemma}
	The consistency of \(\hat{t}\) in the limit of infinite data is shown in the following theorem. 
	\begin{theorem}\label{Tconsistencyt}
		Under the same assumptions as in Theorem \ref{Tconsistencymmd}, we have \(\hat{t}\stackrel{p}{\rightarrow}t\) as \(n_0,n_1\rightarrow\infty\).
	\end{theorem}
	Unfortunately, it is extremely difficult to compute the (asymptotic) null distribution of \(\hat{t}\) analytically, and so we resort to resampling the treatment labels to simulate the null distribution. To ensure that our resampling scheme respects the control and treatment covariate distributions \(P_{X|Z=0}\) and \(P_{X|Z=1}\), we follow the conditional resampling scheme of \citet{rosenbaum1984conditional}. We first estimate the propensity score \(e(x_i)\) for each datapoint \(x_i\) (e.g. using kernel logistic regression (KLR) \cite{zhu2005kernel,marteau2019globally}), and then resample each data label from this estimated propensity score. By repeating this resampling procedure and computing the test statistic on each resampled dataset, we can simulate from the null distribution of the test statistic. Finally, the test statistic computed from the original dataset is compared to this simulated null distribution, and the null hypothesis is rejected or not rejected accordingly. The exact procedure is summarised in Algorithm \ref{Atest}. 
	
	\section{Understanding the CoDiTE}\label{Sunderstanding}
	After determining \textit{whether} \(P_{Y_0|X}\) and \(P_{Y_1|X}\) are different via MMD-associated CoDiTE and hypothesis testing, we now turn to Q2, i.e. we investigate \textit{how} they are different. 
	
	\subsection{Conditional Witness Functions}\label{SSwitness}
	For two real-valued random variables, the witness function between them is a useful tool for visualising where their densities differ, without explicitly estimating the densities (\citealp[Figure 1]{gretton2012kernel}; \citealp[Figure 1]{lloyd2015statistical}). We extend this to the conditional case with the (unnormalised) \textit{conditional witness function} \(\mu_{Y_1|X}-\mu_{Y_0|X}\). 
	
	Let us fix \(x\in\mathcal{X}\). The witness function between \(P_{Y_1|X=x}\) and \(P_{Y_0|X=x}\) is \(\mu_{Y_1|X=x}-\mu_{Y_0|X=x}:\mathcal{Y}\rightarrow\mathbb{R}\). For \(y\in\mathcal{Y}\) in regions where the density of \(P_{Y_1|X=x}\) is greater than that of \(P_{Y_0|X=x}\), we have \(\mu_{Y_1|X=x}(y)-\mu_{Y_0|X=x}(y)>0\). For \(y\) in regions where the converse is true, we similarly have \(\mu_{Y_1|X=x}(y)-\mu_{Y_0|X=x}(y)<0\). The greater the difference in density, the greater the magnitude of the witness function. For each \(y\in\mathcal{Y}\), the associated CoDiTE is
	\[U_{\text{witness},y}(x)=\mu_{Y_1|X=x}(y)-\mu_{Y_0|X=x}(y).\]
	
	The estimates in (\ref{EempiricalF}) can be plugged in to obtain the estimate \(\hat{U}_{\text{witness},y}=\hat{\mu}_{Y_1|X=x}(y)-\hat{\mu}_{Y_0|X=x}(y)\). Since convergence in the RKHS norm implies pointwise convergence \citep[p.10, Corollary 1]{berlinet2004reproducing}, Theorem \ref{Tconsistencymmd} implies the consistency of \(\hat{U}_{\text{witness},y}\) with respect to the corresponding PEHDE. Clearly, if \(X\) is more than 1-dimensional, heat maps as in Figure \ref{Fmotivation}(c) cannot be plotted; however, fixing a particular \(x\in\mathcal{X}\), \(\hat{\mu}_{Y_1|X=x}-\hat{\mu}_{Y_0|X=x}\) can be plotted against \(y\), since \(Y\subseteq\mathbb{R}\). Such plots will be informative of where the density of \(P_{Y_1|X=x}\) is greater than that of \(P_{Y_0|X=x}\) and vice versa. 
	
	\subsection{CoDiTE associated with Specific Distributional Quantities via U-statistic Regression}\label{SScoditemoments}
	
	Next, we consider CoDiTE on specific distributional quantities, such as the mean, variance or skewness, or some function thereof. For example, \citet[Eqn. (2)]{briseno2020flexible} were interested, in addition to the CATE, in the treatment effect on the standard deviation \(U_D(x)=\text{std}(Y_1|X=x)-\text{std}(Y_0|X=x)\). Our motivating example in Figure \ref{Fmotivation} could inspire a \say{standardised} version of the CATE\footnote{In practice, if the CoDiTE involves ratios of estimated quantities, we do not recommend plugging in the estimates directly into the ratio, since, if the denominator is small, then a small error in the estimation of the denominator will result in a large error in the overall CoDiTE estimation. Instead, we recommend that the practitioner estimate the numerator and the denominator separately and interpret the results directly from the raw estimates.}:
	\begin{equation}\label{EstandardisedCATE}
		U_D(x)=\frac{\mathbb{E}[Y_1|X=x]-\mathbb{E}[Y_0|X=x]}{\sqrt{\text{Var}(Y_1|X=x)+\text{Var}(Y_0|X=x)}}.
	\end{equation}
	Many of these quantities can be represented as the expectation of a U-kernel, i.e. \(\mathbb{E}[h(Y_1,...,Y_r)]\) (c.f. Section \ref{SSustatistics}). For example, \(h(y)=y\) gives the mean, \(h(y_1,y_2)=\frac{1}{2}(y_1-y_2)^2\) gives the variance and \(h(y_1,y_2)=\lvert y_1-y_2\rvert\) gives Gini's mean difference. We consider their conditional counterparts, i.e. \(\theta(P_{Y_0|X})=\mathbb{E}[h(Y_{01},...,Y_{0r})|X_1,...,X_r]\) and \(\theta(P_{Y_1|X})=\mathbb{E}[h(Y_{11},...,Y_{1r})|X_1,...,X_r]\) (c.f. Section \ref{SSustatistics}). By \citet[p.146, Theorem 1.17]{cinlar2011probability}, there exist functions \(F_0,F_1:\mathcal{X}^r\rightarrow\mathbb{R}\) such that \(F_0(X_1,...,X_r)=\theta(P_{Y_0|X})\) and \(F_1(X_1,...,X_r)=\theta(P_{Y_1|X})\). 
	
	\begin{table*}[t]
		\caption{Root mean square error in estimating the conditional standard deviation, with standard error from 100 simulations, for GAMLSS (implemented via the R package \texttt{gamlss} \cite{rigby2005generalized}) and our U-statistic regression via generalised kernel ridge regression (U-regression KRR; implemented via the Falkon library on Python \cite{rudi2017falkon,meanti2020kernel}). Lower is better.}
		\vskip 0.1in
		\centering
		\begin{tabular}{|c|c|c|c|c|c|c|}
			\hline
			\multirow{2}{*}{\bfseries Method} & \multicolumn{2}{|c|}{\bfseries Setting SN} & \multicolumn{2}{|c|}{\bfseries Setting LN} & \multicolumn{2}{|c|}{\bfseries Setting HN} \\\cline{2-7}
			& Control & Treatment & Control & Treatment & Control & Treatment \\ \hline
			GAMLSS & \(0.17\pm0.031\) & \(0.767\pm0.414\) & \(3.3\pm0.55\) & \(15.44\pm8.128\) & \(2.27\pm0.44\) & \(10.91\pm5.42\) \\\hline
			U-regression KRR & \(\mathbf{0.13\pm0.059}\)& \(\mathbf{0.16\pm0.059}\) & \(\mathbf{1.1\pm0.31}\) & \(\mathbf{2.16\pm0.61}\) & \(\mathbf{0.7\pm0.25}\) & \(\mathbf{1.39\pm0.47}\) \\ \hline
		\end{tabular}
		\label{Tabcomparison}
	\end{table*}
	\begin{figure*}[t]
		\centerline{\includegraphics[scale=0.45]{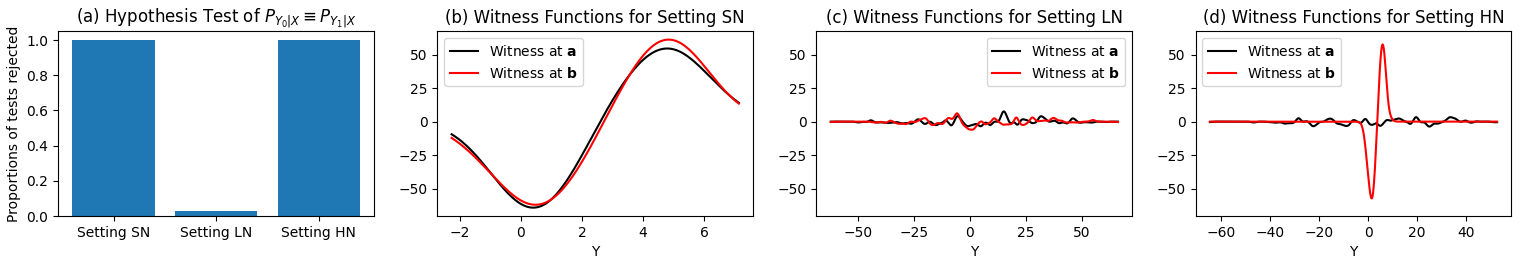}}
		\caption{\textbf{Hypothesis testing and witness functions on the IHDP dataset.} (a) Hypothesis test is conducted on 100 simulations for each setting, with the bar chart showing proportion of tests rejected for each setting. In setting \say{LN}, where the variance overwhelms the CATE, the test does not reject the hypothesis \(P_{Y_0|X}\equiv P_{Y_1|X}\), whereas in the other two settings, the hypothesis is rejected. (b) At both \(X=\mathbf{a}\) and \(X=\mathbf{b}\), the density of the control group is larger than that of the treatment group around \(Y=0\), and the reverse is true around \(Y=4\), showing the marked effect of the treatment. (c) At both \(X=\mathbf{a}\) and \(X=\mathbf{b}\), the density of the control and treatment groups are roughly equal for all \(Y\). (d) At \(X=\mathbf{a}\), where the variance engulfs the CATE, the density of the control and treatment groups are roughly equal for all \(Y\), whereas at \(X=\mathbf{b}\), the witness function clearly shows where the density of one group dominates the other. The juxtaposition of witness functions at different points in the covariate space is an exploratory tool to compare the relative strength of the treatment effect. }
		\label{Fihdp}
	\end{figure*}
	
	Estimation of \(F_0\) and \(F_1\) can be done via U-statistic regression, by generalising kernel ridge regression as follows. As in Section \ref{SSestimationmmd}, let \(k_0:\mathcal{X}\times\mathcal{X}\rightarrow\mathbb{R}\) be a kernel on \(\mathcal{X}\) with RKHS \(\mathcal{H}_0\). Then if we define \(k_0^r:\mathcal{X}^r\times\mathcal{X}^r\rightarrow\mathbb{R}\) as
	\[k^r_0((x_1,...,x_r),(x'_1,...,x'_r))=k_0(x_1,x'_1)...k_0(x_r,x'_r),\]
	\citet[p.31, Theorem 13]{berlinet2004reproducing} tells us that \(k_0^r\) is a reproducing kernel on \(\mathcal{X}^r\) with RKHS \(\mathcal{H}_0^r=\mathcal{H}_0\otimes...\otimes\mathcal{H}_0\), the \(r\)-times tensor product of \(\mathcal{H}_0\), whose elements are functions \(\mathcal{X}^r\rightarrow\mathbb{R}\). We estimate \(F_0\) in \(\mathcal{H}_0^r\). Given any \(F\in\mathcal{H}_0^r\), the natural least-squares risk is
	\[\mathcal{E}(F)=\mathbb{E}[(F(X_1,...,X_r)-h(Y_{01},...,Y_{0r}))^2].\]
	Recalling the control sample \(\{(x^0_i,y^0_i)\}^{n_0}_{i=1}\), we solve the following regularised least-squares problem:
	\begin{equation}\label{Eleastsquares}
		\hat{F}_0=\argmin_{F\in\mathcal{H}_0^r}\left\{\hat{\mathcal{E}}(F)+\lambda^0_{n_0}\left\lVert F\right\rVert_{\mathcal{H}^r_0}^2\right\}
	\end{equation}
	where the empirical least-squares risk \(\hat{\mathcal{E}}\) is defined as
	\[\hat{\mathcal{E}}(F)=\frac{1}{\binom{n_0}{r}}\sum\left(F(x^0_{i_1},...,x^0_{i_r})-h(y^0_{i_1},...,y^0_{i_r})\right)^2,\]
	with the summation over the \(\binom{n_0}{r}\) combinations of \(r\) distinct elements \(\{i_1,...,i_r\}\) from \(\{1,...,n_0\}\). Note that \(\hat{\mathcal{E}}(F)\) is itself a U-statistic for the estimation of \(\mathcal{E}(F)\). The following is a representer theorem for the problem in (\ref{Eleastsquares}). 
	\begin{theorem}\label{Trepresenter}
		The solution \(\hat{F}_0\) to the problem in (\ref{Eleastsquares}) is
		\begin{alignat*}{2}
			&\hat{F}_0(x_1,...,x_r)=\sum^{n_0}_{i_1,...,i_r}k_0(x^0_{i_1},x_1)...k_0(x^0_{i_r},x_r)c^0_{i_1,...,i_r}
		\end{alignat*}
		where the coefficients \(c^0_{i_1,...,i_r}\in\mathbb{R}\) are the unique solution of the \(n^r\) linear equations, 
		\begin{alignat*}{2}
			&\sum^{n_0}_{j_1,...,j_r=1}\left(k_0\left(x^0_{i_1},x^0_{j_1}\right)...k_0\left(x^0_{i_r},x^0_{j_r}\right)\right.\\
			&\left.+\binom{n_0}{r}\lambda^0_{n_0}\delta_{i_1j_1}...\delta_{i_rj_r}\right)c^0_{j_1,...,j_r}=h\left(y^0_{i_1},...,y^0_{i_r}\right).
		\end{alignat*}
	\end{theorem}
	Note that if \(r=1\) and \(h(y)=y\), we recover the usual kernel ridge regression. The following result shows that this estimation procedure is universally consistent. 
	\begin{theorem}\label{Tconsistencyu}
		Suppose \(k_0^r\) is a bounded and universal kernel and that \(\lambda^0_{n_0}\) decays at a slower rate than \(\mathcal{O}(n_0^{-1/2})\). Then as \(n_0\rightarrow\infty\),
		\[\mathbb{E}\left[\left(\hat{F}_0\left(X_1,...,X_r\right)-F_0\left(X_1,...,X_r\right)\right)^2\right]\stackrel{p}{\rightarrow}0.\]
	\end{theorem}
	A consistent estimate \(\hat{F}_1\) of \(F_1\) is obtained by exactly the same procedure, using the treatment sample \(\{(x^1_i,y^1_i)\}^{n_1}_{i=1}\).
	
	\section{Experiments}\label{Sexperiments}
	\subsection{Semi-synthetic IHDP Data}\label{SSihdp}
	We demonstrate the use of our methods on the Infant Health and Development Program (IHDP) dataset \citep[Section 4]{hill2011bayesian}. The covariates are taken from a randomised control trial, from which a non-random portion is removed to imitate an observational study.
	The reason for its popularity in the CATE literature is that, for each datapoint, the outcome is simulated for both treatment and control, enabling cross-validation and evaluation, which is usually not possible in observational studies due to the missing counterfactuals. Existing works first define the noiseless response surfaces for the control and treatment groups, and generate realisations of the potential outcomes by applying Gaussian noise with constant variance across the whole dataset. 
	
	This last assumption of constant variance is somewhat unrealistic, but of little importance in evaluating CATE estimators. In our experiments, we modify the data generating process in three different ways, all of which have the same parallel linear mean response surfaces, with the CATE of 4 (\say{response surface A} in \citet{hill2011bayesian}). In setting \say{SN} (\say{small noise}), the standard deviation of the noise is constant at 1, so that the CATE of 4 translates to a meaningful treatment effect. In setting \say{LN} (\say{large noise}), the standard deviation of the noise is constant at 20, meaning that the mean difference in the response surfaces is negligible in comparison. In this case, our test does not reject the hypothesis that the two conditional distributions are the same, and there is no case for further investigation (see middle bar in Figure \ref{Fihdp}(a)). In setting \say{HN} (\say{heterogeneous noise}), the standard deviation is heterogeneous across the dataset, so that the standard deviation is 1 for some data points while others have standard deviation of 20. The exact data generating process is detailed in Appendix \ref{Sihdpmoredetails}. 
	
	In setting \say{HN}, let us consider points \(\mathbf{a},\mathbf{b}\in\mathcal{X}\) with \(\text{sd}(Y|X=\mathbf{a})=20\) and \(\text{sd}(Y|X=\mathbf{b})=1\). Then even though the CATE at \(\mathbf{a}\) and \(\mathbf{b}\) are equal at 4, we have \(\text{std}(Y_1-Y_0|X=\mathbf{a})\gg\text{std}(Y_1-Y_0|X=\mathbf{b})\), such that there is a pronounced treatment effect at \(\mathbf{b}\), while the variance engulfs the treatment effect at \(\mathbf{a}\). The comparative magnitudes of the witness functions conditioned on \(\mathbf{a}\) and \(\mathbf{b}\) confirm this heterogeneity (see Figure \ref{Fihdp}(d)). In Table \ref{Tabcomparison}, the quality of estimation of the standard deviation via our U-statistic regression is compared with GAMLSS \citep{stasinopoulos2017flexible} estimation for each setting. 
	
	An immediate benefit is a better understanding of the treatment. Even a perfect CATE estimator cannot capture such heterogeneity in distributional treatment effect (variance, in this case). As argued in Section \ref{SSproblem}, any method that involves comparing mean values (of which CATE is one) should also take into account the variance for it to be meaningful. This will give a clearer picture of the subpopulations on which there is a marked treatment effect, and those on which it is weaker, than relying on the CATE alone. Such knowledge should in turn influence policy decisions, in terms of which subpopulations should be targeted. We note that recently \citet{jesson2020identifying} considered CATE uncertainty in IHDP in the context of a different task: making or deferring treatment recommendations while using Bayesian neural networks, focusing on cases where overlap fails or under covariate shift; however, distributional considerations can be important even when overlap is satisfied and no covariate shift takes place. 
	
	\subsection{Real Outcomes: LaLonde Data}\label{SSlalonde}
	\begin{figure}
		\centering
		\includegraphics[scale=0.5]{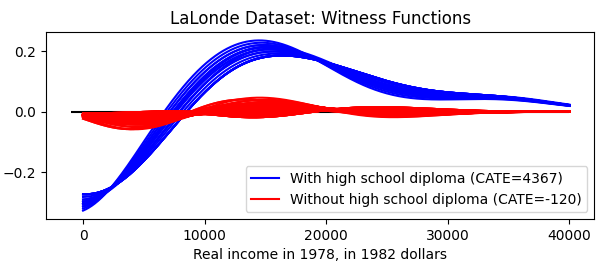}
		\caption{\textbf{Witness functions for Black, unmarried participant up to the age of 25, unemployed in both 1974 and 1975.}
			Each curve (witness function) corresponds to an individual in this subset.}
		\label{Flalonde}
	\end{figure}
	In this section, we apply the proposed methods to LaLonde's well-known National Supported Work (NSW) dataset \citep{LaLonde86:NSW,Dehejia99:LaLonde} which has been used widely to evaluate estimators of treatment effects. The outcome of interest \(Y\) is the real earnings in 1978, with treatment \(Z\) being the job training. We refer the interested readers to \citet[Sec. 2.1]{Dehejia99:LaLonde} for a detailed description of the dataset. As income distributions are known to be skewed to the right, it may be interesting to investigate not only the CATE, but the entire distributions. 
	
	The test rejects the hypothesis \(P_{Y_0|X}\equiv P_{Y_1|X}\) with p-value of 0.013. As a demonstration of the kind of exploratory analysis that can be conducted using the conditional witness functions, we focus our attention on a subset of the data on which the overlap condition is satisfied -- Black, unmarried participants up to the age of 25, who were unemployed in both 1974 and 1975. Figure \ref{Flalonde} shows the witness function for each individual in this subset, with the colour of the curve delineating whether the corresponding individual has a high school diploma. 
	
	We can see clearly that for those without a high school diploma, the treatment effect is not so pronounced, whereas there is a marked treatment effect for those with it. Negative values of the witness function for small income values mean that we are more likely to get small income values from the control group than the treatment group, whereas larger income values are more likely to come from the treatment group, as indicated by the positive values of the witness functions. In particular, the tail of the blue curves to the right implies a skewness of the density of the treated group relative to the control group, and the treatment group continues to have larger density than the control group for high income values (\(>25000\)), albeit to a lesser extent. Such comparison of densities in different regions of \(\mathcal{Y}\) is not possible with the CATE, which is a simple difference of the means between the control and treated groups. 
	
	\section{Discussion \& Conclusion}\label{Sconclusion}
	
	In this paper, we discussed the analysis of the conditional distributional treatment effect (CoDiTE). We first propose a new kernel-based hypothesis test via kernel conditional mean embeddings to see whether there exists any CoDiTE. Then we proceeded to investigate the nature of the treatment effect via conditional witness functions, revealing where and how much the conditional densities differ, and U-statistic regression, which is informative about the differences in specific conditional distributional quantities. 
	
	We foresee that much of the work that has been done by the machine learning community on treatment effect analysis, although cast mostly in the context of CATE, applies for the CoDiTE. Examples include \textit{meta learners} \cite{kunzel2019metalearners}, model validation \cite{alaa2019validating}, subgroup analysis \cite{su2009subgroup,lee2020robust} and covariate balancing \cite{gretton2009covariate,kallus2018optimal}.
	overo
	A major obstacle in any covariate-conditional analysis of treatment effect is this: when the covariate space is high-dimensional, the accuracy and reliability of the estimates deteriorate significantly due to the curse of dimensionality, and we heavily rely on changes to be smooth across the covariate space. This limitation is present not only in methods presented in this paper, but any CATE or CoDiTE analysis. While out of scope for the present paper, it is of interest to investigate how to mitigate this problem.
	
	Last but not least, we argue that the conditional distributional treatment effect can play an important role in making fair and explainable decisions as it provides a more complete picture of the treatment effect.
	On the one hand, policymakers can use tools that we develop to identify the groups of individuals for which the outcome distributions differ most through the effect modifiers.
	On the other hand, the presence of effect modification that is associated with sensitive attributes such as race, ethnicity, and gender creates challenges for decision makers. If they knew that there is effect modification by race, for example, certain groups of individuals may be treated unfairly. In practice, our tools can potentially be used to detect the discrepancy between outcome distributions conditioned on these sensitive attributes, which is also an interesting avenue for future work. 
	
	\section*{Acknowledgements}
	We are very grateful to Jonas K\"ubler at the MPI for Intelligent Systems, T\"ubingen, for readily providing help with running the codes, and for other insightful discussions. We also thank Alexis Derumigny at Delft University of Technology and Giacomo Meanti at Universit\`a degli Studi di Genova for readily answering queries via email and helpful discussions. Finally, we thank Arthur Gretton for his comments on the early draft of our manuscript.
	
	This work was in part supported by the German Federal Ministry of Education and Research (BMBF): Tübingen AI Center, FKZ: 01IS18039B, and by the Machine Learning Cluster of Excellence, EXC number 2064/1 – Project number 390727645. Uri Shalit was supported by the Israel Science Foundation (grant No. 1950/19).

	\bibliography{ref}

\begin{thebibliography}{80}
\providecommand{\natexlab}[1]{#1}
\providecommand{\url}[1]{\texttt{#1}}
\expandafter\ifx\csname urlstyle\endcsname\relax
  \providecommand{\doi}[1]{doi: #1}\else
  \providecommand{\doi}{doi: \begingroup \urlstyle{rm}\Url}\fi

\bibitem[Abadie(2002)]{abadie2002bootstrap}
Abadie, A.
\newblock Bootstrap {T}ests for {D}istributional {T}reatment {E}ffects in
  {I}nstrumental {V}ariable {M}odels.
\newblock \emph{Journal of the American statistical Association}, 97\penalty0
  (457):\penalty0 284--292, 2002.

\bibitem[Alaa \& Schaar(2018)Alaa and Schaar]{alaa2018limits}
Alaa, A. and Schaar, M.
\newblock Limits of {E}stimating {H}eterogeneous {T}reatment {E}ffects:
  {G}uidelines for {P}ractical {A}lgorithm {D}esign.
\newblock In \emph{International Conference on Machine Learning}, pp.\
  129--138, 2018.

\bibitem[Alaa \& Van Der~Schaar(2019)Alaa and Van
  Der~Schaar]{alaa2019validating}
Alaa, A. and Van Der~Schaar, M.
\newblock Validating {C}ausal {I}nference {M}odels via {I}nfluence {F}unctions.
\newblock In \emph{International Conference on Machine Learning}, pp.\
  191--201, 2019.

\bibitem[Alaa \& van~der Schaar(2017)Alaa and van~der Schaar]{alaa2017bayesian}
Alaa, A.~M. and van~der Schaar, M.
\newblock Bayesian {I}nference of {I}ndividualized {T}reatment {E}ffects using
  {M}ulti-{T}ask {G}aussian {P}rocesses.
\newblock In \emph{Advances in Neural Information Processing Systems}, pp.\
  3424--3432, 2017.

\bibitem[Alaa \& van~der Schaar(2018)Alaa and van~der Schaar]{alaa2018bayesian}
Alaa, A.~M. and van~der Schaar, M.
\newblock Bayesian {N}onparametric {C}ausal {I}nference: {I}nformation {R}ates
  and {L}earning {A}lgorithms.
\newblock \emph{IEEE Journal of Selected Topics in Signal Processing},
  12\penalty0 (5):\penalty0 1031--1046, 2018.

\bibitem[Aronszajn(1950)]{aronszajn1950theory}
Aronszajn, N.
\newblock Theory of {R}eproducing {K}ernels.
\newblock \emph{Transactions of the American mathematical society}, 68\penalty0
  (3):\penalty0 337--404, 1950.

\bibitem[Atan et~al.(2018)Atan, Jordon, and van~der Schaar]{atan2018deep}
Atan, O., Jordon, J., and van~der Schaar, M.
\newblock Deep-{T}reat: {L}earning {O}ptimal {P}ersonalized {T}reatments from
  {O}bservational {D}ata using {N}eural {N}etworks.
\newblock In \emph{AAAI}, pp.\  2071--2078, 2018.

\bibitem[Berlinet \& Thomas-Agnan(2004)Berlinet and
  Thomas-Agnan]{berlinet2004reproducing}
Berlinet, A. and Thomas-Agnan, C.
\newblock \emph{Reproducing {K}ernel {H}ilbert {S}paces in {P}robability and
  {S}tatistics}.
\newblock Kluwer Academic Publishers, 2004.

\bibitem[Bitler et~al.(2017)Bitler, Gelbach, and Hoynes]{bitler2017can}
Bitler, M.~P., Gelbach, J.~B., and Hoynes, H.~W.
\newblock Can {V}ariation in {S}ubgroups' {A}verage {T}reatment {E}ffects
  {E}xplain {T}reatment {E}ffect {H}eterogeneity? {E}vidence from a {S}ocial
  {E}xperiment.
\newblock \emph{Review of Economics and Statistics}, 99\penalty0 (4):\penalty0
  683--697, 2017.

\bibitem[Brise{\~n}o~Sanchez et~al.(2020)Brise{\~n}o~Sanchez, Hohberg, Groll,
  and Kneib]{briseno2020flexible}
Brise{\~n}o~Sanchez, G., Hohberg, M., Groll, A., and Kneib, T.
\newblock Flexible {I}nstrumental {V}ariable {D}istributional {R}egression.
\newblock \emph{Journal of the Royal Statistical Society: Series A (Statistics
  in Society)}, 183\penalty0 (4):\penalty0 1553--1574, 2020.

\bibitem[Carmeli et~al.(2006)Carmeli, De~Vito, and Toigo]{carmeli2006vector}
Carmeli, C., De~Vito, E., and Toigo, A.
\newblock Vector {V}alued {R}eproducing {K}ernel {H}ilbert {S}paces of
  {I}ntegrable {F}unctions and {M}ercer {T}heorem.
\newblock \emph{Analysis and Applications}, 4\penalty0 (04):\penalty0 377--408,
  2006.

\bibitem[Carmeli et~al.(2010)Carmeli, De~Vito, Toigo, and
  Umanit{\'a}]{carmeli2010vector}
Carmeli, C., De~Vito, E., Toigo, A., and Umanit{\'a}, V.
\newblock Vector valued reproducing kernel hilbert spaces and universality.
\newblock \emph{Analysis and Applications}, 8\penalty0 (01):\penalty0 19--61,
  2010.

\bibitem[Chang et~al.(2015)Chang, Lee, and Whang]{chang2015nonparametric}
Chang, M., Lee, S., and Whang, Y.-J.
\newblock Nonparametric {T}ests of {C}onditional {T}reatment {E}ffects with an
  {A}pplication to {S}ingle-{S}ex {S}chooling on {A}cademic {A}chievements.
\newblock \emph{The Econometrics Journal}, 18\penalty0 (3):\penalty0 307--346,
  2015.

\bibitem[Chernozhukov et~al.(2013)Chernozhukov, Fern{\'a}ndez-Val, and
  Melly]{chernozhukov2013inference}
Chernozhukov, V., Fern{\'a}ndez-Val, I., and Melly, B.
\newblock Inference on {C}ounterfactual {D}istributions.
\newblock \emph{Econometrica}, 81\penalty0 (6):\penalty0 2205--2268, 2013.

\bibitem[Chernozhukov et~al.(2020)Chernozhukov, Fernandez-Val, and
  Weidner]{chernozhukov2020network}
Chernozhukov, V., Fernandez-Val, I., and Weidner, M.
\newblock Network and {P}anel {Q}uantile {E}ffects via {D}istribution
  {R}egression.
\newblock \emph{Journal of Econometrics}, 2020.

\bibitem[{\c{C}}{\i}nlar(2011)]{cinlar2011probability}
{\c{C}}{\i}nlar, E.
\newblock \emph{Probability and {S}tochastics}, volume 261.
\newblock Springer Science \& Business Media, 2011.

\bibitem[Crump et~al.(2008)Crump, Hotz, Imbens, and
  Mitnik]{crump2008nonparametric}
Crump, R.~K., Hotz, V.~J., Imbens, G.~W., and Mitnik, O.~A.
\newblock Nonparametric {T}ests for {T}reatment {E}ffect {H}eterogeneity.
\newblock \emph{The Review of Economics and Statistics}, 90\penalty0
  (3):\penalty0 389--405, 2008.

\bibitem[Dehejia \& Wahba(1999)Dehejia and Wahba]{Dehejia99:LaLonde}
Dehejia, R.~H. and Wahba, S.
\newblock Causal effects in nonexperimental studies: Reevaluating the
  evaluation of training programs.
\newblock \emph{Journal of the American Statistical Association}, 94\penalty0
  (448):\penalty0 1053--1062, 1999.

\bibitem[Derumigny(2019)]{derumigny2019estimation}
Derumigny, A.
\newblock Estimation of a {R}egular {C}onditional {F}unctional by {C}onditional
  {U}-{S}tatistics {R}egression.
\newblock \emph{arXiv preprint arXiv:1903.10914}, 2019.

\bibitem[Dinculeanu(2000)]{dinculeanu2000vector}
Dinculeanu, N.
\newblock \emph{Vector {I}ntegration and {S}tochastic {I}ntegration in {B}anach
  {S}paces}, volume~48.
\newblock John Wiley \& Sons, 2000.

\bibitem[Foster et~al.(2011)Foster, Taylor, and Ruberg]{foster2011subgroup}
Foster, J.~C., Taylor, J.~M., and Ruberg, S.~J.
\newblock Subgroup {I}dentification from {R}andomized {C}linical {T}rial
  {D}ata.
\newblock \emph{Statistics in medicine}, 30\penalty0 (24):\penalty0 2867--2880,
  2011.

\bibitem[Fukumizu et~al.(2008)Fukumizu, Gretton, Sun, and
  Sch{\"o}lkopf]{fukumizu2008kernel}
Fukumizu, K., Gretton, A., Sun, X., and Sch{\"o}lkopf, B.
\newblock Kernel {M}easures of {C}onditional {D}ependence.
\newblock In \emph{Advances in neural information processing systems}, pp.\
  489--496, 2008.

\bibitem[Fukumizu et~al.(2013)Fukumizu, Song, and Gretton]{fukumizu2013kernel}
Fukumizu, K., Song, L., and Gretton, A.
\newblock Kernel {B}ayes' {R}ule: {B}ayesian {I}nference with {P}ositive
  {D}efinite {K}ernels.
\newblock \emph{The Journal of Machine Learning Research}, 14\penalty0
  (1):\penalty0 3753--3783, 2013.

\bibitem[Gretton et~al.(2009)Gretton, Smola, Huang, Schmittfull, Borgwardt, and
  Sch{\"o}lkopf]{gretton2009covariate}
Gretton, A., Smola, A., Huang, J., Schmittfull, M., Borgwardt, K., and
  Sch{\"o}lkopf, B.
\newblock Covariate {S}hift by {K}ernel {M}ean {M}atching.
\newblock \emph{Dataset shift in machine learning}, 3\penalty0 (4):\penalty0 5,
  2009.

\bibitem[Gretton et~al.(2012)Gretton, Borgwardt, Rasch, Sch{\"o}lkopf, and
  Smola]{gretton2012kernel}
Gretton, A., Borgwardt, K.~M., Rasch, M.~J., Sch{\"o}lkopf, B., and Smola, A.
\newblock A {K}ernel {T}wo-{S}ample {T}est.
\newblock \emph{Journal of Machine Learning Research}, 13\penalty0
  (Mar):\penalty0 723--773, 2012.

\bibitem[Hahn et~al.(2020)Hahn, Murray, and Carvalho]{hahn2020bayesian}
Hahn, P.~R., Murray, J.~S., and Carvalho, C.~M.
\newblock Bayeisan {R}egression {T}ree {M}odels for {C}ausal {I}nference:
  {R}egularisation, {C}onfounding, and {H}eterogeneous {E}ffects (with
  {D}iscussion).
\newblock \emph{Bayesian Analysis}, 15\penalty0 (3):\penalty0 965--1056, 09
  2020.

\bibitem[Hill(2011)]{hill2011bayesian}
Hill, J.~L.
\newblock Bayesian {N}onparametric {M}odeling for {C}ausal {I}nference.
\newblock \emph{Journal of Computational and Graphical Statistics}, 20\penalty0
  (1):\penalty0 217--240, 2011.

\bibitem[Hoeffding(1948)]{hoeffding1948class}
Hoeffding, W.
\newblock A {C}lass of {S}tatistics with {A}symptotically {N}ormal
  {D}istribution.
\newblock \emph{The Annals of Mathematical Statistics}, pp.\  293--325, 1948.

\bibitem[Hohberg et~al.(2020)Hohberg, P{\"u}tz, and
  Kneib]{hohberg2020treatment}
Hohberg, M., P{\"u}tz, P., and Kneib, T.
\newblock Treatment {E}ffects {B}eyond the {M}ean {U}sing {D}istributional
  {R}egression: {M}ethods and {G}uidance.
\newblock \emph{Plos one}, 15\penalty0 (2):\penalty0 e0226514, 2020.

\bibitem[Holland(1986)]{holland1986statistics}
Holland, P.~W.
\newblock Statistics and {C}ausal {I}nference.
\newblock \emph{Journal of the American statistical Association}, 81\penalty0
  (396):\penalty0 945--960, 1986.

\bibitem[Imbens \& Rubin(2015)Imbens and Rubin]{imbens2015causal}
Imbens, G.~W. and Rubin, D.~B.
\newblock \emph{Causal {I}nference in {S}tatistics, {S}ocial, and {B}iomedical
  sciences}.
\newblock Cambridge University Press, 2015.

\bibitem[Imbens \& Wooldridge(2009)Imbens and Wooldridge]{imbens2009recent}
Imbens, G.~W. and Wooldridge, J.~M.
\newblock Recent {D}evelopments in the {E}conometrics of {P}rogram
  {E}valuation.
\newblock \emph{Journal of economic literature}, 47\penalty0 (1):\penalty0
  5--86, 2009.

\bibitem[Jesson et~al.(2020)Jesson, Mindermann, Shalit, and
  Gal]{jesson2020identifying}
Jesson, A., Mindermann, S., Shalit, U., and Gal, Y.
\newblock Identifying {C}ausal-{E}ffect {I}nference {F}ailure with
  {U}ncertainty-{A}ware {M}odels.
\newblock \emph{Advances in Neural Information Processing Systems}, 33, 2020.

\bibitem[Johansson et~al.(2016)Johansson, Shalit, and
  Sontag]{johansson2016learning}
Johansson, F., Shalit, U., and Sontag, D.
\newblock Learning {R}epresentations for {C}ounterfactual {I}nference.
\newblock In \emph{International conference on machine learning}, pp.\
  3020--3029, 2016.

\bibitem[Kadri et~al.(2016)Kadri, Duflos, Preux, Canu, Rakotomamonjy, and
  Audiffren]{kadri2016operator}
Kadri, H., Duflos, E., Preux, P., Canu, S., Rakotomamonjy, A., and Audiffren,
  J.
\newblock Operator-{V}alued {K}ernels for {L}earning from {F}unctional
  {R}esponse {D}ata.
\newblock \emph{The Journal of Machine Learning Research}, 17\penalty0
  (1):\penalty0 613--666, 2016.

\bibitem[Kallus(2018)]{kallus2018optimal}
Kallus, N.
\newblock Optimal {A} {P}riori {B}alance in the {D}esign of {C}ontrolled
  {E}xperiments.
\newblock \emph{Journal of the Royal Statistical Society Series B}, 80\penalty0
  (1):\penalty0 85--112, 2018.

\bibitem[Kim et~al.(2018)Kim, Kim, and Kennedy]{kim2018causal}
Kim, K., Kim, J., and Kennedy, E.~H.
\newblock Causal {E}ffects {B}ased on {D}istributional {D}istances.
\newblock \emph{arXiv preprint arXiv:1806.02935}, 2018.

\bibitem[Koenker(2005)]{koenker2005quantile}
Koenker, R.
\newblock \emph{Quantile {R}egression}.
\newblock Cambridge University Press, 2005.

\bibitem[K{\"u}nzel et~al.(2019)K{\"u}nzel, Sekhon, Bickel, and
  Yu]{kunzel2019metalearners}
K{\"u}nzel, S.~R., Sekhon, J.~S., Bickel, P.~J., and Yu, B.
\newblock Metalearners for {E}stimating {H}eterogeneous {T}reatment {E}ffects
  using {M}achine {L}earning.
\newblock \emph{Proceedings of the national academy of sciences}, 116\penalty0
  (10):\penalty0 4156--4165, 2019.

\bibitem[LaLonde(1986)]{LaLonde86:NSW}
LaLonde, R.~J.
\newblock Evaluating the econometric evaluations of training programs with
  experimental data.
\newblock \emph{The American Economic Review}, 76\penalty0 (4):\penalty0
  604--620, 1986.

\bibitem[Lee et~al.(2020)Lee, Zhang, Zame, Shen, Lee, and van~der
  Schaar]{lee2020robust}
Lee, H.-S., Zhang, Y., Zame, W., Shen, C., Lee, J.-W., and van~der Schaar, M.
\newblock Robust {R}ecursive {P}artitioning for {H}eterogeneous {T}reatment
  {E}ffects with {U}ncertainty {Q}uantification.
\newblock \emph{Advances in Neural Information Processing Systems}, 33, 2020.

\bibitem[Lee(2009)]{lee2009non}
Lee, M.-J.
\newblock Non-parametric {T}ests for {D}istributional {T}reatment {E}ffect for
  {R}andomly {C}ensored {R}esponses.
\newblock \emph{Journal of the Royal Statistical Society: Series B (Statistical
  Methodology)}, 71\penalty0 (1):\penalty0 243--264, 2009.

\bibitem[Lee \& Whang(2009)Lee and Whang]{lee2009nonparametric}
Lee, S.~S. and Whang, Y.-J.
\newblock Nonparametric {T}ests of {C}onditional {T}reatment {E}ffects.
\newblock Technical report, Cowles Foundation for Research in Economics, Yale
  University, 2009.

\bibitem[Lloyd \& Ghahramani(2015)Lloyd and Ghahramani]{lloyd2015statistical}
Lloyd, J.~R. and Ghahramani, Z.
\newblock Statistical {M}odel {C}riticism using {K}ernel {T}wo {S}ample
  {T}ests.
\newblock \emph{Advances in Neural Information Processing Systems},
  28:\penalty0 829--837, 2015.

\bibitem[Louizos et~al.(2017)Louizos, Shalit, Mooij, Sontag, Zemel, and
  Welling]{louizos2017causal}
Louizos, C., Shalit, U., Mooij, J.~M., Sontag, D., Zemel, R., and Welling, M.
\newblock Causal {E}ffect {I}nference with {D}eep {L}atent-{V}ariable {M}odels.
\newblock In \emph{Advances in Neural Information Processing Systems}, pp.\
  6446--6456, 2017.

\bibitem[Marteau-Ferey et~al.(2019)Marteau-Ferey, Bach, and
  Rudi]{marteau2019globally}
Marteau-Ferey, U., Bach, F., and Rudi, A.
\newblock Globally {C}onvergent {N}ewton {M}ethods for {I}ll-{C}onditioned
  {G}eneralized {S}elf-{C}oncordant {L}osses.
\newblock In \emph{Advances in Neural Information Processing Systems}, 2019.

\bibitem[Meanti et~al.(2020)Meanti, Carratino, Rosasco, and
  Rudi]{meanti2020kernel}
Meanti, G., Carratino, L., Rosasco, L., and Rudi, A.
\newblock Kernel {M}ethods {T}hrough the {R}oof: {H}andling {B}illions of
  {P}oints {E}fficiently.
\newblock \emph{Advances in Neural Information Processing Systems}, 33, 2020.

\bibitem[Micchelli \& Pontil(2005)Micchelli and Pontil]{micchelli2005learning}
Micchelli, C.~A. and Pontil, M.
\newblock On {L}earning {V}ector-{V}alued {F}unctions.
\newblock \emph{Neural computation}, 17\penalty0 (1):\penalty0 177--204, 2005.

\bibitem[Muandet et~al.(2017)Muandet, Fukumizu, Sriperumbudur, Sch{\"o}lkopf,
  et~al.]{muandet2017kernel}
Muandet, K., Fukumizu, K., Sriperumbudur, B., Sch{\"o}lkopf, B., et~al.
\newblock Kernel {M}ean {E}mbedding of {D}istributions: {A} {R}eview and
  {B}eyond.
\newblock \emph{Foundations and Trends{\textregistered} in Machine Learning},
  10\penalty0 (1-2):\penalty0 1--141, 2017.

\bibitem[Muandet et~al.(2018)Muandet, Kanagawa, Saengkyongam, and
  Marukatat]{muandet2020counterfactual}
Muandet, K., Kanagawa, M., Saengkyongam, S., and Marukatat, S.
\newblock Counterfactual {M}ean {E}mbedding.
\newblock \emph{arXiv preprint arXiv:1805.08845}, 2018.

\bibitem[Nadaraya(1964)]{nadaraya1964estimating}
Nadaraya, E.~A.
\newblock On {E}stimating {R}egression.
\newblock \emph{Theory of Probability \& Its Applications}, 9\penalty0
  (1):\penalty0 141--142, 1964.

\bibitem[Park \& Muandet(2020{\natexlab{a}})Park and Muandet]{park2020measure}
Park, J. and Muandet, K.
\newblock A {M}easure-{T}heoretic {A}pproach to {K}ernel {C}onditional {M}ean
  {E}mbeddings.
\newblock In \emph{Advances in Neural Information Processing Systems},
  2020{\natexlab{a}}.

\bibitem[Park \& Muandet(2020{\natexlab{b}})Park and
  Muandet]{park2020regularised}
Park, J. and Muandet, K.
\newblock Regularised {L}east-{S}quares {R}egression with
  {I}nfinite-{D}imensional {O}utput {S}pace.
\newblock \emph{arXiv preprint arXiv:2010.10973}, 2020{\natexlab{b}}.

\bibitem[Powers et~al.(2018)Powers, Qian, Jung, Schuler, Shah, Hastie, and
  Tibshirani]{powers2018some}
Powers, S., Qian, J., Jung, K., Schuler, A., Shah, N.~H., Hastie, T., and
  Tibshirani, R.
\newblock Some {M}ethods for {H}eterogeneous {T}reatment {E}ffect {E}stimation
  in {H}igh {D}imensions.
\newblock \emph{Statistics in medicine}, 37\penalty0 (11):\penalty0 1767--1787,
  2018.

\bibitem[Rigby \& Stasinopoulos(2005)Rigby and
  Stasinopoulos]{rigby2005generalized}
Rigby, R.~A. and Stasinopoulos, D.~M.
\newblock Generalized {A}dditive {M}odels for {L}ocation, {S}cale and
  {S}hape,(with discussion).
\newblock \emph{Applied Statistics}, 54:\penalty0 507--554, 2005.

\bibitem[Rosenbaum(1984)]{rosenbaum1984conditional}
Rosenbaum, P.~R.
\newblock Conditional {P}ermutation {T}ests and the {P}ropensity {S}core in
  {O}bservational {S}tudies.
\newblock \emph{Journal of the American Statistical Association}, 79\penalty0
  (387):\penalty0 565--574, 1984.

\bibitem[Rosenbaum(2002)]{rosenbaum2002observational}
Rosenbaum, P.~R.
\newblock \emph{Observational {S}tudies}.
\newblock Springer Science \& Business Media, 2002.

\bibitem[Rosenbaum \& Rubin(1983)Rosenbaum and Rubin]{rosenbaum1983central}
Rosenbaum, P.~R. and Rubin, D.~B.
\newblock The {C}entral {R}ole of the {P}ropensity {S}core in {O}bservational
  {S}tudies for {C}ausal {E}ffects.
\newblock \emph{Biometrika}, 70\penalty0 (1):\penalty0 41--55, 1983.

\bibitem[Rubin(2005)]{rubin2005causal}
Rubin, D.~B.
\newblock Causal {I}nference using {P}otential {O}utcomes: {D}esign,
  {M}odeling, {D}ecisions.
\newblock \emph{Journal of the American Statistical Association}, 100\penalty0
  (469):\penalty0 322--331, 2005.

\bibitem[Rudi et~al.(2017)Rudi, Carratino, and Rosasco]{rudi2017falkon}
Rudi, A., Carratino, L., and Rosasco, L.
\newblock Falkon: {A}n {O}ptimal {L}arge {S}cale {K}ernel {M}ethod.
\newblock In \emph{Advances in Neural Information Processing Systems}, pp.\
  3888--3898, 2017.

\bibitem[Scholkopf \& Smola(2001)Scholkopf and Smola]{scholkopf2001learning}
Scholkopf, B. and Smola, A.~J.
\newblock \emph{Learning with {K}ernels: {S}upport {V}ector {M}achines,
  {R}egularization, {O}ptimization, and {B}eyond}.
\newblock MIT press, 2001.

\bibitem[Schwabik \& Ye(2005)Schwabik and Ye]{schwabik2005topics}
Schwabik, S. and Ye, G.
\newblock \emph{Topics in {B}anach {S}pace {I}ntegration}, volume~10.
\newblock World Scientific, 2005.

\bibitem[Serfling(1980)]{serfling1980approximation}
Serfling, R.~J.
\newblock \emph{Approximation {T}heorems of {M}athematical {S}tatistics}.
\newblock John Wiley \& Sons, 1980.

\bibitem[Shalit et~al.(2017)Shalit, Johansson, and
  Sontag]{shalit2017estimating}
Shalit, U., Johansson, F.~D., and Sontag, D.
\newblock Estimating {I}ndividual {T}reatment {E}ffect: {G}eneralization
  {B}ounds and {A}lgorithms.
\newblock In \emph{International Conference on Machine Learning}, pp.\
  3076--3085. PMLR, 2017.

\bibitem[Shen(2019)]{shen2019estimation}
Shen, S.
\newblock Estimation and {I}nference of {D}istributional {P}artial {E}ffects:
  {T}heory and {A}pplication.
\newblock \emph{Journal of Business \& Economic Statistics}, 37\penalty0
  (1):\penalty0 54--66, 2019.

\bibitem[Shi et~al.(2019)Shi, Blei, and Veitch]{shi2019adapting}
Shi, C., Blei, D., and Veitch, V.
\newblock Adapting {N}eural {N}etworks for the {E}stimation of {T}reatment
  {E}ffects.
\newblock In \emph{Advances in Neural Information Processing Systems}, pp.\
  2507--2517, 2019.

\bibitem[Simon-Gabriel \& Sch{\"o}lkopf(2018)Simon-Gabriel and
  Sch{\"o}lkopf]{simongabriel2018kernel}
Simon-Gabriel, C.-J. and Sch{\"o}lkopf, B.
\newblock Kernel {D}istribution {E}mbeddings: {U}niversal {K}ernels,
  {C}haracteristic {K}ernels and {K}ernel {M}etrics on {D}istributions.
\newblock \emph{The Journal of Machine Learning Research}, 19\penalty0
  (1):\penalty0 1708--1736, 2018.

\bibitem[Singh et~al.(2020)Singh, Xu, and Gretton]{singh2020kernel}
Singh, R., Xu, L., and Gretton, A.
\newblock Kernel {M}ethods for {P}olicy {E}valuation: {T}reatment {E}ffects,
  {M}ediation {A}nalysis, and {O}ff-{P}olicy {P}lanning.
\newblock \emph{arXiv preprint arXiv:2010.04855}, 2020.

\bibitem[Smola et~al.(2007)Smola, Gretton, Song, and
  Sch{\"o}lkopf]{smola2007hilbert}
Smola, A., Gretton, A., Song, L., and Sch{\"o}lkopf, B.
\newblock A {H}ilbert {S}pace {E}mbedding for {D}istributions.
\newblock In \emph{International Conference on Algorithmic Learning Theory},
  pp.\  13--31. Springer, 2007.

\bibitem[Song et~al.(2009)Song, Huang, Smola, and Fukumizu]{song2009hilbert}
Song, L., Huang, J., Smola, A., and Fukumizu, K.
\newblock Hilbert {S}pace {E}mbeddings of {C}onditional {D}istributions with
  {A}pplications to {D}ynamical {S}ystems.
\newblock In \emph{Proceedings of the 26th Annual International Conference on
  Machine Learning}, pp.\  961--968, 2009.

\bibitem[Song et~al.(2013)Song, Fukumizu, and Gretton]{song2013kernel}
Song, L., Fukumizu, K., and Gretton, A.
\newblock Kernel {E}mbeddings of {C}onditional {D}istributions: {A} {U}nified
  {K}ernel {F}ramework for {N}onparametric {I}nference in {G}raphical {M}odels.
\newblock \emph{IEEE Signal Processing Magazine}, 30\penalty0 (4):\penalty0
  98--111, 2013.

\bibitem[Sriperumbudur et~al.(2010)Sriperumbudur, Gretton, Fukumizu,
  Sch{\"o}lkopf, and Lanckriet]{sriperumbudur2010hilbert}
Sriperumbudur, B.~K., Gretton, A., Fukumizu, K., Sch{\"o}lkopf, B., and
  Lanckriet, G.~R.
\newblock Hilbert {S}pace {E}mbeddings and {M}etrics on {P}robability
  {M}easures.
\newblock \emph{Journal of Machine Learning Research}, 11\penalty0
  (Apr):\penalty0 1517--1561, 2010.

\bibitem[Sriperumbudur et~al.(2011)Sriperumbudur, Fukumizu, and
  Lanckriet]{sriperumbudur2011universality}
Sriperumbudur, B.~K., Fukumizu, K., and Lanckriet, G.~R.
\newblock Universality, {C}haracteristic {K}ernels and {RKHS} {E}mbedding of
  {M}easures.
\newblock \emph{Journal of Machine Learning Research}, 12\penalty0
  (Jul):\penalty0 2389--2410, 2011.

\bibitem[Stasinopoulos et~al.(2017)Stasinopoulos, Rigby, Heller, Voudouris, and
  De~Bastiani]{stasinopoulos2017flexible}
Stasinopoulos, M.~D., Rigby, R.~A., Heller, G.~Z., Voudouris, V., and
  De~Bastiani, F.
\newblock \emph{Flexible {R}egression and {S}moothing: {U}sing {GAMLSS} in
  {R}}.
\newblock CRC Press, 2017.

\bibitem[Stute(1991)]{stute1991conditional}
Stute, W.
\newblock Conditional {U}-{S}tatistics.
\newblock \emph{The Annals of Probability}, 19\penalty0 (2):\penalty0 812--825,
  1991.

\bibitem[Su et~al.(2009)Su, Tsai, Wang, Nickerson, and Li]{su2009subgroup}
Su, X., Tsai, C.-L., Wang, H., Nickerson, D.~M., and Li, B.
\newblock Subgroup {A}nalysis via {R}ecursive {P}artitioning.
\newblock \emph{Journal of Machine Learning Research}, 10\penalty0 (2), 2009.

\bibitem[Wager \& Athey(2018)Wager and Athey]{wager2018estimation}
Wager, S. and Athey, S.
\newblock Estimation and {I}nference of {H}eterogeneous {T}reatment {E}ffects
  using {R}andom {F}orests.
\newblock \emph{Journal of the American Statistical Association}, 113\penalty0
  (523):\penalty0 1228--1242, 2018.

\bibitem[Watson(1964)]{watson1964smooth}
Watson, G.~S.
\newblock Smooth {R}egression {A}nalysis.
\newblock \emph{Sankhy{\=a}: The Indian Journal of Statistics, Series A}, pp.\
  359--372, 1964.

\bibitem[Yoon et~al.(2018)Yoon, Jordon, and van~der Schaar]{yoon2018ganite}
Yoon, J., Jordon, J., and van~der Schaar, M.
\newblock G{ANITE}: {E}stimation of {I}ndividualized {T}reatment {E}ffects
  using {G}enerative {A}dversarial {N}ets.
\newblock In \emph{International Conference on Learning Representations}, 2018.

\bibitem[Zhu \& Hastie(2005)Zhu and Hastie]{zhu2005kernel}
Zhu, J. and Hastie, T.
\newblock Kernel {L}ogistic {R}egression and the {I}mport {V}ector {M}achine.
\newblock \emph{Journal of Computational and Graphical Statistics}, 14\penalty0
  (1):\penalty0 185--205, 2005.

\end{thebibliography}
	\bibliographystyle{icml2021}
	\appendix
	\onecolumn
	\section{Background Material}\label{Smoredetails}
	In this section, we give a more detailed review of the background on reproducing kernel Hilbert space embeddings and U-statistics. Interested readers can refer to \citet{berlinet2004reproducing,muandet2017kernel} for the former, and \citet[Chapter 5]{serfling1980approximation} for the latter. 
	\subsection{Reproducing Kernel Hilbert Space Embeddings}
	Let \(\mathcal{H}\) be a vector space of real-valued functions on \(\mathcal{Y}\), endowed with the structure of a Hilbert space via an inner product \(\langle\cdot,\cdot\rangle_\mathcal{H}\). Let \(\lVert\cdot\rVert_\mathcal{H}\) be the associated norm, i.e. \(\lVert f\rVert_\mathcal{H}=\langle f,f\rangle_\mathcal{H}^{\frac{1}{2}}\) for \(f\in\mathcal{H}\).
	\begin{definition}[{\citet[p.7, Definition 1]{berlinet2004reproducing}}]
		A function \(l:\mathcal{Y}\times\mathcal{Y}\rightarrow\mathbb{R}\) is a \textit{reproducing kernel} of the Hilbert space \(\mathcal{H}\) if and only if
		\begin{enumerate}[(i)]
			\item for all \(y\in\mathcal{Y}\), \(l(y,\cdot)\in\mathcal{H}\); 
			\item for all \(y\in\mathcal{Y}\) and for all \(f\in\mathcal{H}\), \(\langle f,l(y,\cdot)\rangle_\mathcal{H}=f(y)\) (the \textit{reproducing property}).
		\end{enumerate}
		A Hilbert space of functions \(\mathcal{Y}\rightarrow\mathbb{R}\) which possesses a reproducing kernel is called the \textit{reproducing kernel Hilbert space} (RKHS). 
	\end{definition}
	For any \(y\in\mathcal{Y}\), denote by \(e_y:\mathcal{H}\rightarrow\mathbb{R}\) the evaluation functional at \(y\), i.e. \(e_y(f)=f(y)\) for \(f\in\mathcal{H}\). Riesz representation theorem can be used to prove the following lemma. 
	\begin{lemma}[{\citet[p.9, Theorem 1]{berlinet2004reproducing}}]
		A Hilbert space of functions \(\mathcal{Y}\rightarrow\mathbb{R}\) has a reproducing kernel if and only if all evaluation functionals \(e_y,y\in\mathcal{Y}\) are continuous on \(\mathcal{H}\).
	\end{lemma}
	Next, we characterise reproducing kernels. 
	\begin{definition}[{\citet[p.10, Definition 2]{berlinet2004reproducing}}]
		A function \(l:\mathcal{Y}\times\mathcal{Y}\rightarrow\mathbb{R}\) is called a \textit{positive definite function} if, for all \(n\geq1\), any \(a_1,...,a_n\in\mathbb{R}\) and any \(y_1,...,y_n\in\mathcal{Y}\), 
		\[\sum^n_{i,j=1}a_ia_jl(y_i,y_j)\geq0.\]
	\end{definition}
	A reproducing kernel is a positive definite function, since, by the reproducing property,
	\[\sum^n_{i,j=1}a_ia_jl(y_i,y_j)=\left\lVert\sum^n_{i=1}a_il(y_i,\cdot)\right\rVert^2_\mathcal{H}\geq0\]
	(see \citet[p.13, Lemma 2]{berlinet2004reproducing}). The Moore-Aronszajn Theorem \citep{aronszajn1950theory} shows that the set of positive definite functions and the set of reproducing kernels on \(\mathcal{Y}\times\mathcal{Y}\) are identical. 
	\begin{theorem}[{\citet[p.19, Theorem 3]{berlinet2004reproducing}}]
		Let \(l\) be a positive definite function on \(\mathcal{Y}\times\mathcal{Y}\). Then there exists a unique Hilbert space of functions \(\mathcal{Y}\rightarrow\mathbb{R}\) with \(l\) as its reproducing kernel. The subspace \(\tilde{\mathcal{H}}\) of \(\mathcal{H}\) spanned by \(\{l(y,\cdot):y\in\mathcal{Y}\}\) is dense in \(\mathcal{H}\), and \(\mathcal{H}\) is the set of functions \(\mathcal{Y}\rightarrow\mathbb{R}\) which are pointwise limits of Cauchy sequences in \(\tilde{\mathcal{H}}\) with the inner product
		\[\langle f,g\rangle_{\tilde{\mathcal{H}}}=\sum^n_{i=1}\sum^m_{j=1}\alpha_i\beta_jl(y_i,y_j)\]
		where \(f=\sum^n_{i=1}\alpha_il(y_i,\cdot)\) and \(g=\sum^m_{j=1}\beta_jl(y_j,\cdot)\). 
	\end{theorem}
	Examples of commonly used kernels in Euclidean spaces include the linear kernel \(l(y,y')=y\cdot y'\), the monomial kernel \(l(y,y')=(y\cdot y')^p\), the polynomial kernel \(l(y,y')=(y\cdot y'+1)^p\), the Gaussian kernel \(l(y,y')=e^{-\frac{1}{\sigma^2}\lVert y-y'\rVert_2^2}\) and the Laplacian kernel \(l(y,y')=e^{-\frac{1}{\sigma^2}\lVert y-y'\rVert_1}\).
	
	Kernel methods in machine learning turns linear methods into non-linear ones using the so-called \say{kernel trick}, whereby individual datapoints \(y\in\mathcal{Y}\) are \say{embedded} into an RKHS \(\mathcal{H}\) with reproducing kernel \(l\) via the mapping \(y\mapsto l(y,\cdot)\). The RKHS is high- (and often infinite-)dimensional, and performing a linear method (e.g. linear regression, support vector machine, principal component analysis, etc.) in \(\mathcal{H}\) with datapoints \(l(y_i,\cdot),i=1,...,n\), instead of the original space \(\mathcal{Y}\) with datapoints \(y_i,i=1,...,n\), results in a nonlinear method in the original space. Please see \citet{scholkopf2001learning} for more details. 
	
	Recently, this idea of RKHS embeddings has been extended to embed entire (conditional) distributions, rather than individual datapoints, via the expectation. Suppose \(Y\) is a random variable taking values in \(\mathcal{Y}\), with distribution \(P_Y\). Assuming the integrability condition \(\int_\mathcal{Y}\sqrt{l(y,y)}dP_Y(y)<\infty\), we define the \textit{kernel mean embedding} \(\mu_{P_Y}\in\mathcal{H}\) of the measure \(P_Y\), or the random variable \(Y\), as
	\[\mu_{P_Y}(\cdot)=\mathbb{E}\left[l(Y,\cdot)\right]=\int_\mathcal{Y}l(y,\cdot)dP_Y(y)=\int_\Omega l(Y(\omega),\cdot)dP(\omega).\]
	Note that the integrand \(l(Y,\cdot)\) is an element in a Hilbert space (and therefore a Banach space), so the integral is not the usual Lebesgue integral on \(\mathbb{R}\). There are a number of ways in which one can define integration on a Banach space \citep{schwabik2005topics}. Among those, the Bochner integral \citep[p.15, Definition 35]{dinculeanu2000vector} is the simplest and most intuitive one, and suffices for our purposes. Riesz representation theorem is again used to prove the following mean embedding version of the reproducing property. 
	\begin{lemma}[{\citet{smola2007hilbert}}]
		For each \(f\in\mathcal{Y}\),
		\[\mathbb{E}\left[f(Y)\right]=\int_\mathcal{Y}f(y)dP_{Y}(y)=\left\langle f,\mu_{P_Y}\right\rangle_\mathcal{H}.\]
	\end{lemma}
	Using the kernel mean embedding, we can define a distance function, called the \textit{maximum mean discrepancy} \citep{gretton2012kernel}, between two random variables \(Y\) and \(Y'\) on \(\mathcal{Y}\), or equivalently, two probability measures \(P_Y\) and \(P_{Y'}\), as
	\[\text{MMD}(Y,Y')=\left\lVert\mu_{P_Y}-\mu_{P_{Y'}}\right\rVert_\mathcal{H}.\]
	The name maximum mean discrepancy comes from the following lemma. 
	\begin{lemma}[{\citet[Lemma 4]{gretton2012kernel}}]
		We have
		\[\textnormal{MMD}(Y,Y')=\sup_{f\in\mathcal{H},\left\lVert f\right\rVert_\mathcal{H}\leq1}\left\{\mathbb{E}\left[f(Y)\right]-\mathbb{E}\left[f(Y')\right]\right\}.\]
	\end{lemma}
	In this alternative definition of the MMD, the function in the unit ball of \(\mathcal{H}\) that maximises \(\mathbb{E}[f(Y)]-\mathbb{E}[f(Y')]\) is called the \textit{witness function} \citep[Section 2.3]{gretton2012kernel}. It can easily be seen that the witness function is in fact
	\[\frac{\mu_{P_Y}-\mu_{P_{Y'}}}{\left\lVert\mu_{P_Y}-\mu_{P_{Y'}}\right\rVert_\mathcal{H}}.\]
	\citet{lloyd2015statistical} uses the unnormalised witness function \(\mu_{P_Y}-\mu_{P_{Y'}}\) for model criticism. 
	
	The MMD is not a proper metric, since \(Y\) and \(Y'\) may be distinct and still give \(\text{MMD}(Y,Y')=0\), depending on the kernel \(l\) that is used. The notion of \textit{characteristic kernels} is therefore essential, since it tells us whether the associated RKHS is rich enough to enable us to distinguish distinct distributions based on their embeddings. 
	\begin{definition}[{\citet[Section 2.2]{fukumizu2008kernel}}]
		Denote by \(\mathcal{P}\) the set of all probability measures on \(\mathcal{Y}\). A positive definite kernel \(l\) is \textit{characteristic} if the kernel mean embedding map \(\mathcal{P}\rightarrow\mathcal{H}:P_Y\mapsto\mu_{P_Y}\) is injective.
	\end{definition}
	For example, of the aforementioned kernels, the Gaussian and Laplacian kernels are characteristic, whereas the linear, monomial and polynomial kernels are not. The MMD associated with a characteristic kernel is then a proper metric between probability measures on \(\mathcal{Y}\). See \citet{sriperumbudur2010hilbert,sriperumbudur2011universality,simongabriel2018kernel} for various characterisations of characteristic kernels. 
	
	Now we discuss conditional embedding of distributions into RKHSs. Suppose \(X\) is a random variable on a space \(\mathcal{X}\). 
	\begin{definition}[{\citet[Definition 3.1]{park2020measure}}]
		The \textit{conditional mean embedding} of the random variable \(Y\), or equivalently, the distribution \(P_Y\), is the Bochner conditional expectation (as defined in \citet[p.45, Definition 38]{dinculeanu2000vector})
		\[\mu_{P_{Y|X}}=\mathbb{E}\left[l(Y,\cdot)\mid X\right].\]
	\end{definition}
	Notice that this is a straightforward extension of the kernel mean embedding \(\mu_{P_Y}=\mathbb{E}[l(Y,\cdot)]\) to the conditional case. 
	
	\subsection{U-Statistics}
	Suppose \(Y_1,Y_2,...,Y_r\) are independent copies of the random variable \(Y\), i.e. they are independent and all have distribution \(P_Y\). Let \(h:\mathcal{Y}^r\rightarrow\mathbb{R}\) be a symmetric function (called a \textit{kernel} in the U-statistics literature; confusion must be avoided with the reproducing kernel used throughout this paper), i.e. for any permutation \(\pi\) of \(\{1,...,r\}\), we have \(h(y_1,...,y_r)=h(y_{\pi(1)},...,y_{\pi(r)})\). Suppose we would like to estimate a function of the form
	\[\theta(P_Y)=\mathbb{E}\left[h\left(Y_1,...,Y_r\right)\right]=\int_\mathcal{Y}...\int_\mathcal{Y}h\left(y_1,...,y_r\right)dP_Y(y_1)...dP_Y(y_r).\]
	The corresponding \textit{U-statistic} for an unbiased estimation of \(\theta(P_Y)\) based on a sample \(Y_1,...,Y_n\) of size \(n\geq r\) is given by
	\[\hat{\theta}(P_Y)=\frac{1}{\binom{n}{r}}\sum h\left(Y_{i_1},...,Y_{i_r}\right),\]
	where \(\binom{n}{r}\) is the binomial coefficient and the summation is over the \(\binom{n}{r}\) combinations of \(r\) distinct elements \(\{i_1,...,i_r\}\) from \(\{1,...,n\}\). Clearly, since the expectation of each summand yields \(\theta(P_Y)\), we have \(\mathbb{E}[\hat{\theta}(P_Y)]=\theta(P_Y)\), so U-statistics are unbiased estimators. 
	
	Some examples of \(h\) and the corresponding estimator include the sample mean \(h(y)=y\), the sample variance \(h(y_1,y_2)=\frac{1}{2}(y_1-y_2)^2\), the sample cumulative distribution up to \(y^*\) \(h(y)=\mathbf{1}(y\leq y^*)\), the \(k^\text{th}\) sample raw moment \(h(y)=y^k\) and Gini's mean difference \(h(y_1,y_2)=\lvert y_1-y_2\rvert\). 
	
	To the best of our knowledge, \citet{stute1991conditional} was the first to consider a conditional counterpart of U-statistics. Let \(X_1,...,X_r\) be independent copies of the random variable \(X\). We are now interested in the estimation of the following quantity:
	\[\theta\left(P_{Y|X}\right)=\mathbb{E}\left[h\left(Y_1,...,Y_r\right)\mid X_1,...,X_r\right].\]
	By \citet[p.146, Theorem 1.17]{cinlar2011probability}, \(\theta(P_{Y|X})\) can be considered as a function \(\mathcal{X}^r\rightarrow\mathbb{R}\), such that for each \(r\)-tuple \(\{x_1,...,x_r\}\), we have
	\[\theta\left(P_{Y|X}\right)\left(x_1,...,x_r\right)=\mathbb{E}\left[h\left(Y_1,...,Y_r\right)\mid X_1=x_1,...,X_r=x_r\right].\]
	The simplest case is when \(r=1\) and \(h(y)=y\). In this case, the estimand reduces to \(f(X)=\mathbb{E}[Y|X]\), which is the usual regression problem for which a plethora of methods exist. Suppose we have a sample \(\{(X_i,Y_i)\}_{i=1}^n\). One such regression method is the Nadaraya-Watson kernel smoother:
	\[\hat{f}(x)=\frac{\sum^n_{i=1}Y_iK\left(\frac{x-X_i}{a}\right)}{\sum^n_{i=1}K\left(\frac{x-X_i}{a}\right)},\]
	where \(K\) is the so-called \say{smoothing kernel} and \(a\) is the bandwidth. This was extended by \citet{stute1991conditional} to \(r\geq1\) and more general \(h\):
	\[\hat{\theta}\left(P_{Y|X}\right)\left(x_1,...,x_r\right)=\frac{\sum h\left(Y_{i_1},...,Y_{i_r}\right)\prod^r_{j=1}K\left(\frac{x_j-X_{i_j}}{a}\right)}{\sum\prod^r_{j=1}K\left(\frac{x_j-X_{i_j}}{a}\right)},\]
	where the sums are over the \(\binom{n}{r}\) combinations of \(r\) distinct elements \(\{i_1,...,i_r\}\) from \(\{1,...,n\}\) as before. \citet{derumigny2019estimation} considers a parametric model of the form
	\[\Lambda\left(\theta\left(P_{Y|X}\right)\left(x_1,...,x_r\right)\right)=\boldsymbol{\psi}\left(x_1,...,x_r\right)^T\beta^*,\]
	where \(\Lambda\) is a strictly increasing and continuously differentiable \say{link function} such that the range of \(\Lambda\circ\theta\) is exactly \(\mathbb{R}\), \(\beta^*\in\mathbb{R}^s\) is the true parameter and \(\boldsymbol{\psi}(\cdot)=\left(\psi_1(\cdot),...,\psi_s(\cdot)\right)^T\in\mathbb{R}^s\) is some basis, such as polynomials, exponentials, indicator functions etc. However, the estimation of \(\beta^*\) still makes use of the Nadaraya-Watson kernel smoothers considered above. 
	
	Of course, Nadaraya-Watson kernel smoothers are far from being the only method of regression that can be extended to estimate conditional U-statistics, and in the main body of the paper (Section \ref{SScoditemoments}), we consider extending kernel ridge regression for this purpose. 
	
	\section{More Details on IHDP Dataset}\label{Sihdpmoredetails}
	In this section, we give more details on the data generating process of the semi-synthetic IHDP (Infant Health and Development Program) dataset that was first used in the treatment effect literature by \citet{hill2011bayesian}. 
	
	The data consists of 25 covariates: birth weight, head circumference, weeks born preterm, birth order, first born, neonatal health index, sex, twin status, whether or not the mother smoked during pregnancy, whether or not the mother drank alcohol during pregnancy, whether or not the mother took drugs during pregnancy, the mother's age, marital status, education attainment, whether or not the mother worked during pregnancy, whether she received prenatal care, and 7 dummy variables for the 8 sites in which the family resided at the start of the intervention. 
	
	These covariates are originally taken from a randomised experiment, and included information about the ethnicity of the mothers. \citet{hill2011bayesian} removed all children with nonwhite mothers from the treatment group, which is clearly a non-random (biased) portion of the data, thereby imitating an observational study. This leaves 608 children in the control group and 139 in the treatment group. The overlap condition is now only satisfied for the treatment group. 
	
	In creating the parallel linear response surfaces, which are used in all three of the settings \say{SN}, \say{LN} and \say{HN}, we let \(\mathbb{E}[Y_0|X]=\beta X\) and \(\mathbb{E}[Y_1|X]=\beta X+4\), where the 25-dimensional coefficient vector \(\beta\) is generated in the same way as in \citet{alaa2018limits}: for the 6 continuous variables (birth weight, head circumference, weeks born preterm, birth order, neonatal health index, mother's age), the corresponding coefficients is sampled from \(\{0,0.1,0.2,0.3,0.4\}\) with probabilities \(\{0.5,0.125,0.125,0.125,0.125\}\) respectively, whereas for the other 19 binary variables, the corresponding coefficients are sampled from \(\{0,0.1,0.2,0.3,0.4\}\) with probabilities \(\{0.6,0.1,0.1,0.1,0.1\}\) respectively. 
	
	Finally, we generate realisations of the potential outcomes by adding noise to the mean response surfaces. We let \(Y_0=\beta X+\epsilon(X)\) and \(Y_1=\beta X+4+\epsilon(X)\), where \(\epsilon(X)=\epsilon_\text{SN}\) in setting \say{SN}, \(\epsilon(X)=\epsilon_\text{LN}\) in setting \say{LN} and \(\epsilon(X)=X_6\epsilon_\text{SN}+(1-X_6)\epsilon_\text{LN}\) in setting \say{HN}, with \(\epsilon_\text{SN}\sim\mathcal{N}(0,1^2)\) and \(\epsilon_\text{LN}\sim\mathcal{N}(0,20^2)\). The covariate \(X_6\) corresponds to the sex of the child, and was chosen because there are roughly the same number of each sex in both the control and the treatment groups. 
	
	\section{Proofs}\label{Sproofs}
	\begin{customproof}{Lemma 4.1}
		For each \(x\in\mathcal{X}\), we have
		\[\hat{U}_\textnormal{MMD}^2(x)=\bm{k}_0^T(x)\mathbf{W}_0\mathbf{L}_0\mathbf{W}_0^T\bm{k}_0(x)-2\bm{k}^T_0(x)\mathbf{W}_0\mathbf{L}\mathbf{W}^T_1\bm{k}_1(x)+\bm{k}_1^T(x)\mathbf{W}_1\mathbf{L}_1\mathbf{W}_1^T\bm{k}_1(x),\]
		where \([\mathbf{L}_0]_{1\leq i,j\leq n_0}=l(y^0_i,y^0_j)\), \([\mathbf{L}]_{1\leq i\leq n_0,1\leq j\leq n_1}=l(y^0_i,y^1_j)\) and \([\mathbf{L}_1]_{1\leq i,j\leq n_1}=l(y^1_i,y^1_j)\).
	\end{customproof}
	\begin{proof}
		We use the reproducing property of \(\mathcal{H}\) and (\ref{EempiricalF}) to see that, for any \(x\in\mathcal{X}\), 
		\begin{alignat*}{2}
			\hat{U}_\textnormal{MMD}^2(x)&=\left\lVert\hat{\mu}_{Y_1|X=x}-\hat{\mu}_{Y_0|X=x}\right\rVert_\mathcal{H}^2\\
			&=\left\lVert\bm{k}_0^T(x)\mathbf{W}_0\bm{l}_0-\bm{k}_1^T(x)\mathbf{W}_1\bm{l}_1\right\rVert_\mathcal{H}^2\\
			&=\left\langle\sum^{n_0}_{i,j=1}k_0(x,x^0_i)\mathbf{W}_{0,ij}l(y^0_j,\cdot),\sum^{n_0}_{p,q=1}k_0(x,x^0_p)\mathbf{W}_{0,pq}l(y^0_q,\cdot)\right\rangle_\mathcal{H}\\
			&\quad-2\left\langle\sum^{n_0}_{i,j=1}k_0(x,x^0_i)\mathbf{W}_{0,ij}l(y^0_j,\cdot),\sum^{n_1}_{p,q=1}k_1(x,x^1_p)\mathbf{W}_{1,pq}l(y^1_q,\cdot)\right\rangle_\mathcal{H}\\
			&\qquad+\left\langle\sum^{n_1}_{i,j=1}k_1(x,x^1_i)\mathbf{W}_{1,ij}l(y^1_j,\cdot),\sum^{n_1}_{p,q=1}k_1(x,x^1_p)\mathbf{W}_{1,pq}l(y^1_q,\cdot)\right\rangle_\mathcal{H}\\
			&=\sum^{n_0}_{i,j,p,q=1}k_0(x,x^0_i)\mathbf{W}_{0,ij}l(y^0_j,y^0_q)\mathbf{W}_{0,qp}^Tk_0(x^0_p,x)\\
			&\quad-2\sum^{n_0}_{i,j=1}\sum^{n_1}_{p,q=1}k_0(x,x^0_i)\mathbf{W}_{0,ij}l(y^0_j,y^1_q)\mathbf{W}_{1,qp}^Tk_1(x^1_p,x)\\
			&\qquad+\sum^{n_1}_{i,j,p,q=1}k_1(x,x^1_i)\mathbf{W}_{1,ij}l(y^1_j,y^1_q)\mathbf{W}_{1,qp}^Tk_1(x^1_p,x)\\
			&=\bm{k}_0^T(x)\mathbf{W}_0\mathbf{L}_0\mathbf{W}_0^T\bm{k}_0(x)-2\bm{k}^T_0(x)\mathbf{W}_0\mathbf{L}\mathbf{W}^T_1\bm{k}_1(x)+\bm{k}_1^T(x)\mathbf{W}_1\mathbf{L}_1\mathbf{W}_1^T\bm{k}_1(x).
		\end{alignat*}
	\end{proof}
	\begin{customproof}{Theorem 4.2}
		Suppose that \(k_0,k_1\) and \(l\) are bounded, that \(\Gamma_0\) and \(\Gamma_1\) are universal, and that \(\lambda^0_{n_0}\) and \(\lambda^1_{n_1}\) decay at slower rates than \(\mathcal{O}(n_0^{-1/2})\) and \(\mathcal{O}(n_1^{-1/2})\) respectively. Then as  \(n_0,n_1\rightarrow\infty\), 
		\[\psi_\textnormal{MMD}\left(\hat{U}_\textnormal{MMD}\right)=\mathbb{E}\left[\left(\hat{U}_\textnormal{MMD}(X)-U_\textnormal{MMD}(X)\right)^2\right]\stackrel{P}{\rightarrow}0.\]
	\end{customproof}
	\begin{proof}
		The simple inequality \(\lVert a+b\rVert^2\leq2\lVert a\rVert^2+2\lVert b\rVert^2\) holds in any Hilbert space. Using this, we see that
		\begin{alignat*}{3}
			\psi_\textnormal{MMD}\left(\hat{U}_\textnormal{MMD}\right)&=\mathbb{E}\left[\left(\hat{U}_\textnormal{MMD}(X)-U_\textnormal{MMD}(X)\right)^2\right]\\
			&=\mathbb{E}\left[\left(\left\lVert\hat{\mu}_{Y_1|X}-\hat{\mu}_{Y_0|X}\right\rVert_\mathcal{H}-\left\lVert\mu_{Y_1|X}-\mu_{Y_0|X}\right\rVert_\mathcal{H}\right)^2\right]\\
			&\leq\mathbb{E}\left[\left\lVert\hat{\mu}_{Y_1|X}-\mu_{Y_1|X}-\hat{\mu}_{Y_0|X}+\mu_{Y_0|X}\right\rVert_\mathcal{H}^2\right]&&\text{by the reverse triangle inequality}\\
			&\leq2\mathbb{E}\left[\left\lVert\hat{\mu}_{Y_1|X}-\mu_{Y_1|X}\right\rVert_\mathcal{H}^2+\left\lVert\hat{\mu}_{Y_0|X}-\mu_{Y_0|X}\right\rVert^2_\mathcal{H}\right]&&\text{by the above inequality}.
		\end{alignat*}
		Hence, it suffices to know that
		\[\mathbb{E}\left[\left\lVert\hat{\mu}_{Y_1|X}-\mu_{Y_1|X}\right\rVert_\mathcal{H}^2\right]\stackrel{p}{\rightarrow}0\qquad\text{and}\qquad\mathbb{E}\left[\left\lVert\hat{\mu}_{Y_0|X}-\mu_{Y_0|X}\right\rVert_\mathcal{H}^2\right]\stackrel{p}{\rightarrow}0.\]
		But this follows immediately from \citet{park2020regularised}, so the proof is complete. 
	\end{proof}
	\begin{customproof}{Lemma 4.3}
		If \(l\) is a characteristic kernel, \(P_{Y_0|X}\equiv P_{Y_1|X}\) if and only if \(t=0\). 
	\end{customproof}
	\begin{proof}
		We can assume without loss of generality that \(P_{Y_0|X}\) and \(P_{Y_1|X}\) are obtained from a regular version of \(P(\cdot\mid X)\). Then by \citep[Theorem 2.9]{park2020measure}, there exist \(C_0,C_1\in\mathcal{F}\) with \(P(C_0)=P(C_1)=1\) such that for all \(\omega\in C_0\), \(\mu_{Y_0|X}(\omega)=\int_\mathcal{Y}l(y,\cdot)dP_{Y_0|X}(\omega)(y)\) and for all \(\omega'\in C_1\), \(\mu_{Y_1|X}(\omega')=\int_\mathcal{Y}l(y,\cdot)dP_{Y_1|X}(\omega')(y)\).
		
		Suppose for contradiction that there exists some measurable \(A\subseteq\mathcal{X}\) with \(P_X(A)>0\) such that for all \(x\in A\), \(\mu_{Y_0|X=x}\neq\int_\mathcal{Y}l(y,\cdot)dP_{Y_0|X=x}(y)\). Then \(P(X^{-1}(A))=P_X(A)>0\), and hence \(P(X^{-1}(A)\cap C_0)>0\). For all \(\omega\in X^{-1}(A)\cap C_0\), we have \(X(\omega)\in A\), and hence
		\[\mu_{Y_0|X}(\omega)\neq\int_\mathcal{Y}l(y,\cdot)dP_{Y_0|X=X(\omega)}(y)=\int_\mathcal{Y}l(y,\cdot)P_{Y_0|X}(\omega)(dy)=\mu_{Y_0|X}(\omega).\]
		This is a contradiction, hence there does not exist a measurable \(A\subseteq\mathcal{X}\) with \(P_X(A)>0\) such that for all \(x\in A\), \(\mu_{Y_0|X=x}\neq\int_\mathcal{Y}l(y,\cdot)dP_{Y_0|X=x}(y)\). Therefore, there must exist some measurable \(A_0\subseteq\mathcal{X}\) with \(P_X(A_0)=1\) such that for all \(x\in A_0\), \(\mu_{Y_0|X=x}=\int_\mathcal{Y}l(y,\cdot)dP_{Y_0|X=x}(y)\). Similarly, there must exist some measurable \(A_1\subseteq\mathcal{X}\) with \(P_X(A_1)=1\) such that for all \(x\in A_1\), \(\mu_{Y_1|X=x}=\int_\mathcal{Y}l(y,\cdot)dP_{Y_1|X=x}(y)\). 
		\begin{description}
			\item[(\(\implies\))] Suppose that \(P_{Y_0|X}\equiv P_{Y_1|X}\). This means that there exists a measurable \(A\subseteq\mathcal{X}\) with \(P_X(A)=1\) such that for all \(x\in A\), the measures \(P_{Y_0|X=x}(\cdot)\) and \(P_{Y_1|X=x}(\cdot)\) are the same. Then for all \(x\in A\cap A_0\cap A_1\),
			\begin{alignat*}{3}
				\mu_{Y_0|X=x}&=\int_\mathcal{Y}l(y,\cdot)dP_{Y_0|X=x}(y)\qquad&&\text{since }x\in A_0\\
				&=\int_\mathcal{Y}l(y,\cdot)dP_{Y_1|X=x}(y)&&\text{since }x\in A\\
				&=\mu_{Y_1|X=x}&&\text{since }x\in A_1.
			\end{alignat*}
			Now, we have \(P_X(A)=P_X(A_0)=P_X(A_1)=1\), so \(P_X(A\cap A_0\cap A_1)=1\). Since \(\mu_{Y_0|X=x}=\mu_{Y_1|X=x}\) for all \(x\in A\cap A_0\cap A_1\), we have \(\mu_{Y_0|X=\cdot}=\mu_{Y_1|X=\cdot}\) \(P_X\)-almost everywhere. Hence, 
			\begin{alignat*}{2}
				t=\mathbb{E}\left[\left\lVert\mu_{Y_1|X}-\mu_{Y_0|X}\right\rVert_\mathcal{H}^2\right]=0
			\end{alignat*}
			\item[(\(\impliedby\))] Now suppose that \(t=0\), i.e. \(\mu_{Y_0|X=\cdot}=\mu_{Y_1|X=\cdot}\) \(P_X\)-almost everywhere, say on a measurable set \(A\subseteq\mathcal{X}\) with \(P_X(A)=1\). Suppose \(x\in A\cap A_0\cap A_1\). Then
			\begin{alignat*}{3}
				\int_\mathcal{Y}l(y,\cdot)dP_{Y_0|X=x}(y)&=\mu_{Y_0|X=x}&&\text{since }x\in A_0\\
				&=\mu_{Y_1|X=x}&&\text{since }x\in A\\
				&=\int_\mathcal{Y}l(y,\cdot)dP_{Y_1|X=x}(y)\qquad&&\text{since }x\in A_1.
			\end{alignat*}
			Since \(k_\mathcal{Y}\) is characteristic, this means that \(P_{Y_0|X=x}\) and \(P_{Y_1|X=x}\) are the same measure. As before, we have \(P_X(A\cap A_0\cap A_1)=1\), hence \(P_{Y_0|X}\equiv P_{Y_1|X}\). 
		\end{description}
	\end{proof}
	\begin{customproof}{Lemma 4.4}
		We have
		\[\hat{t}=\frac{1}{n}\textnormal{Tr}\left(\tilde{\mathbf{K}}_0\mathbf{W}_0\mathbf{L}_0\mathbf{W}_0^T\tilde{\mathbf{K}}^T_0\right)-\frac{2}{n}\textnormal{Tr}\left(\tilde{\mathbf{K}}_0\mathbf{W}_0\mathbf{L}\mathbf{W}^T_1\tilde{\mathbf{K}}^T_1\right)+\frac{1}{n}\textnormal{Tr}\left(\tilde{\mathbf{K}}_1\mathbf{W}_1\mathbf{L}_1\mathbf{W}_1^T\tilde{\mathbf{K}}^T_1\right),\]
		where \(\mathbf{L}_0,\mathbf{L}_1\) and \(\mathbf{L}\) are as defined in Lemma \ref{LhatU} and \([\tilde{\mathbf{K}}_0]_{1\leq i\leq n,1\leq j\leq n_0}=k_0(x_i,x^0_j)\) and \([\tilde{\mathbf{K}}_1]_{1\leq i\leq n,1\leq j\leq n_1}=k_1(x_i,x^1_j)\).
	\end{customproof}
	\begin{proof}
		See that, using the reproducing property in \(\mathcal{H}\) again, 
		\begin{alignat*}{2}
			\hat{t}&=\frac{1}{n}\sum^n_{i=1}\left\lVert\hat{\mu}_{Y_1|X=x_i}-\hat{\mu}_{Y_0|X=x_i}\right\rVert^2_\mathcal{H}\\
			&=\frac{1}{n}\sum^n_{i=1}\left\{\left\lVert\hat{\mu}_{Y_1|X=x_i}\right\rVert^2_\mathcal{H}-2\left\langle\hat{\mu}_{Y_1|X=x_i},\hat{\mu}_{Y_0|X=x_i}\right\rangle_\mathcal{H}+\left\lVert\hat{\mu}_{Y_0|X=x_i}\right\rVert^2_\mathcal{H}\right\}\\
			&=\frac{1}{n}\sum^n_{i=1}\left\{\left\lVert\bm{k}^T_0(x_i)\mathbf{W}_0\bm{l}_0\right\rVert^2_\mathcal{H}-2\left\langle\bm{k}^T_0(x_i)\mathbf{W}_0\bm{l}_0,\bm{k}^T_1(x_i)\mathbf{W}_1\bm{l}_1\right\rangle_\mathcal{H}+\left\lVert\bm{k}^T_1(x_i)\mathbf{W}_1\bm{l}_1\right\rVert^2_\mathcal{H}\right\}\\
			&=\frac{1}{n}\sum^n_{i=1}\left\langle\sum^{n_0}_{p,q=1}k_0(x^0_p,x_i)\mathbf{W}_{0,pq}l(y^0_q,\cdot),\sum^{n_0}_{r,s=1}k_0(x^0_r,x_i)\mathbf{W}_{0,rs}l(y^0_s,\cdot)\right\rangle_\mathcal{H}\\
			&\quad-\frac{2}{n}\sum^n_{i=1}\left\langle\sum^{n_0}_{p,q=1}k_0(x^0_p,x_i)\mathbf{W}_{0,pq}l(y^0_q,\cdot),\sum^{n_1}_{r,s=1}k_1(x^1_r,x_i)\mathbf{W}_{1,rs}l(y^1_s,\cdot)\right\rangle_\mathcal{H}\\
			&\qquad+\frac{1}{n}\sum^n_{i=1}\left\langle\sum^{n_1}_{p,q=1}k_1(x^1_p,x_i)\mathbf{W}_{1,pq}l(y^1_q,\cdot),\sum^{n_1}_{r,s=1}k_1(x^1_r,x_i)\mathbf{W}_{1,rs}l(y^1_s,\cdot)\right\rangle_\mathcal{H}\\
			&=\frac{1}{n}\sum^n_{i=1}\sum^{n_0}_{p,q,r,s=1}k_0(x_i,x^0_p)\mathbf{W}_{0,pq}l(y^0_q,y^0_s)\mathbf{W}^T_{0,sr}k_0(x^0_r,x_i)\\
			&\quad-\frac{2}{n}\sum^n_{i=1}\sum^{n_0}_{p,q=1}\sum^{n_1}_{r,s=1}k_0(x_i,x^0_p)\mathbf{W}_{0,pq}l(y^0_q,y^1_s)\mathbf{W}^T_{1,sr}k_1(x^1_r,x_i)\\
			&\qquad+\frac{1}{n}\sum^n_{i=1}\sum^{n_1}_{p,q,r,s=1}k_1(x_i,x^1_p)\mathbf{W}_{1,pq}l(y^1_q,y^1_s)\mathbf{W}^T_{1,sr}k_1(x^1_r,x_i)\\
			&=\frac{1}{n}\left\{\textnormal{Tr}\left(\tilde{\mathbf{K}}_0\mathbf{W}_0\mathbf{L}_0\mathbf{W}_0^T\tilde{\mathbf{K}}^T_0\right)-2\textnormal{Tr}\left(\tilde{\mathbf{K}}_0\mathbf{W}_0\mathbf{L}\mathbf{W}^T_1\tilde{\mathbf{K}}^T_1\right)+\textnormal{Tr}\left(\tilde{\mathbf{K}}_1\mathbf{W}_1\mathbf{L}_1\mathbf{W}_1^T\tilde{\mathbf{K}}^T_1\right)\right\}
		\end{alignat*}
	\end{proof}
	\begin{customproof}{Theorem 4.5}
		Under the same assumptions as in Theorem \ref{Tconsistencymmd}, we have \(\hat{t}\stackrel{p}{\rightarrow}t\) as \(n_0,n_1\rightarrow\infty\).
	\end{customproof}
	\begin{proof}
		We decompose \(\left\lvert\hat{t}-t\right\rvert\) as follows using the triangle inequality:
		\begin{alignat*}{2}
			\left\lvert\hat{t}-t\right\rvert&=\left\lvert\frac{1}{n}\sum^n_{i=1}\left\lVert\hat{\mu}_{Y_1|X=x_i}-\hat{\mu}_{Y_0|X=x_i}\right\rVert_\mathcal{H}^2-\mathbb{E}\left[\left\lVert\mu_{Y_1|X}-\mu_{Y_0|X}\right\rVert_\mathcal{H}^2\right]\right\rvert\\
			&\leq\left\lvert\frac{1}{n}\sum^n_{i=1}\left\lVert\hat{\mu}_{Y_1|X=x_i}-\hat{\mu}_{Y_0|X=x_i}\right\rVert_\mathcal{H}^2-\mathbb{E}\left[\left\lVert\hat{\mu}_{Y_1|X}-\hat{\mu}_{Y_0|X}\right\rVert^2_\mathcal{H}\right]\right\rvert\\
			&\qquad+\left\lvert\mathbb{E}\left[\left\lVert\hat{\mu}_{Y_1|X}-\hat{\mu}_{Y_0|X}\right\rVert^2_\mathcal{H}\right]-\mathbb{E}\left[\left\lVert\mu_{Y_1|X}-\mu_{Y_0|X}\right\rVert_\mathcal{H}^2\right]\right\rvert
		\end{alignat*}
		Here, the first term converges to 0 in probability by the uniform law of large numbers. For the second term, see that
		\begin{alignat*}{2}
			&\left\lvert\mathbb{E}\left[\left\lVert\hat{\mu}_{Y_1|X}-\hat{\mu}_{Y_0|X}\right\rVert^2_\mathcal{H}\right]-\mathbb{E}\left[\left\lVert\mu_{Y_1|X}-\mu_{Y_0|X}\right\rVert_\mathcal{H}^2\right]\right\rvert\\
			&=\left\lvert\mathbb{E}\left[\left\lVert\hat{\mu}_{Y_1|X}-\mu_{Y_1|X}+\mu_{Y_1|X}-\mu_{Y_0|X}+\mu_{Y_0|X}-\hat{\mu}_{Y_0|X}\right\rVert^2_\mathcal{H}-\left\lVert\mu_{Y_1|X}-\mu_{Y_0|X}\right\rVert_\mathcal{H}^2\right]\right\rvert\\
			&=\left\lvert\mathbb{E}\left[\left\lVert\hat{\mu}_{Y_1|X}-\mu_{Y_1|X}\right\rVert^2_\mathcal{H}+\left\lVert\mu_{Y_0|X}-\hat{\mu}_{Y_0|X}\right\rVert^2_\mathcal{H}\right]+2\left\langle\hat{\mu}_{Y_1|X}-\mu_{Y_1|X},\mu_{Y_1|X}-\mu_{Y_0|X}\right\rangle_\mathcal{H}\right.\\
			&\qquad+\left.2\left\langle\hat{\mu}_{Y_0|X}-\mu_{Y_0|X},\mu_{Y_1|X}-\mu_{Y_0|X}\right\rangle_\mathcal{H}+2\left\langle\hat{\mu}_{Y_1|X}-\mu_{Y_1|X},\hat{\mu}_{Y_0|X}-\hat{\mu}_{Y_0|X}\right\rangle_\mathcal{H}\right\rvert.
		\end{alignat*}
		Here, we have
		\[\mathbb{E}\left[\left\lVert\hat{\mu}_{Y_1|X}-\mu_{Y_1|X}\right\rVert_\mathcal{H}^2\right]\stackrel{p}{\rightarrow}0\qquad\text{and}\qquad\mathbb{E}\left[\left\lVert\hat{\mu}_{Y_0|X}-\mu_{Y_0|X}\right\rVert_\mathcal{H}^2\right]\stackrel{p}{\rightarrow}0\]
		as in the proof of Theorem \ref{Tconsistencymmd}, so we are done. 
	\end{proof}
	\begin{customproof}{Theorem 5.1}
		The solution \(\hat{F}_0\) to the problem in (\ref{Eleastsquares}) is
		\[\hat{F}_0\left(x_1,...,x_r\right)=\sum^{n_0}_{i_1,...,i_r=1}k_0\left(x^0_{i_1},x_1\right)...k_0\left(x^0_{i_r},x_r\right)c_{i_1,...,i_r}\]
		where the coefficients \(c_{i_1,...,i_r}\in\mathbb{R}\) are the unique solution of the \(n^r\) linear equations
		\[\sum^{n_0}_{j_1,...,j_r=1}\left(k_0\left(x^0_{i_1},x^0_{j_1}\right)...k_0\left(x^0_{i_r},x^0_{j_r}\right)+\binom{n_0}{r}\lambda^0_{n_0}\delta_{i_1j_1}...\delta_{i_rj_r}\right)c_{j_1,...,j_r}=h\left(y^0_{i_1},...,y^0_{i_r}\right).\]
	\end{customproof}
	\begin{proof}
		Recall from (\ref{Eleastsquares}) that
		\[\hat{F}_0=\argmin_{F\in\mathcal{H}_0^r}\left\{\frac{1}{\binom{n_0}{r}}\sum\left(F\left(x^0_{i_1},...,x^0_{i_r}\right)-h\left(y^0_{i_1},...,y^0_{i_r}\right)\right)^2+\lambda^0_{n_0}\left\lVert F\right\rVert_{\mathcal{H}^r_0}^2\right\},\]
		where the summation is over the \(\binom{n_0}{r}\) combinations of \(r\) distinct elements \(\{i_1,...,i_r\}\) from \(1,...,n_0\). Write
		\[\hat{F}'_0\left(x_1,...,x_r\right)=\sum^{n_0}_{i_1,...,i_r=1}k_0\left(x^0_{i_1},x_1\right)...k_0\left(x^0_{i_r},x_r\right)c_{i_1,...,i_r}\]
		where the coefficients \(c_{i_1,...,i_r}\in\mathbb{R}\) are the unique solution of the \(n^r\) linear equations
		\[\sum^{n_0}_{j_1,...,j_r=1}\left(k_0\left(x^0_{i_1},x^0_{j_1}\right)...k_0\left(x^0_{i_r},x^0_{j_r}\right)+\binom{n_0}{r}\lambda^0_{n_0}\delta_{i_1j_1}...\delta_{i_rj_r}\right)c_{j_1,...,j_r}=h\left(y^0_{i_1},...,y^0_{i_r}\right).\]
		Also, for any \(F\in\mathcal{H}^r_0\), write \(\hat{\mathcal{E}}_\text{reg}(F)\) for the empirical regularised least-squares risk of \(F\):
		\[\hat{\mathcal{E}}_\text{reg}(F)=\frac{1}{\binom{n_0}{r}}\sum\left(F\left(x^0_{i_1},...,x^0_{i_r}\right)-h\left(y^0_{i_1},...,y^0_{i_r}\right)\right)^2+\lambda^0_{n_0}\left\lVert F\right\rVert_{\mathcal{H}^r_0}^2,\]
		so that \(\hat{F}_0=\argmin_{F\in\mathcal{H}_0^r}\hat{\mathcal{E}}_\text{reg}(F)\). We will show that \(\hat{F}'_0=\hat{F}_0\). For any \(F\in\mathcal{H}^r_0\), write \(G=F-\hat{F}'_0\). Then
		\begin{alignat*}{2}
			\hat{\mathcal{E}}_\text{reg}(F)&=\frac{1}{\binom{n_0}{r}}\sum\left(F\left(x^0_{i_1},...,x^0_{i_r}\right)-h\left(y^0_{i_1},...,y^0_{i_r}\right)\right)^2+\lambda^0_{n_0}\left\lVert F\right\rVert_{\mathcal{H}^r_0}^2\\
			&=\frac{1}{\binom{n_0}{r}}\sum\left(F\left(x^0_{i_1},...,x^0_{i_r}\right)-\hat{F}'_0\left(x^0_{i_1},...,x^0_{i_r}\right)+\hat{F}'_0\left(x^0_{i_1},...,x^0_{i_r}\right)-h\left(y^0_{i_1},...,y^0_{i_r}\right)\right)^2+\lambda^0_{n_0}\left\lVert F\right\rVert_{\mathcal{H}^r_0}^2\\
			&=\hat{\mathcal{E}}_\text{reg}\left(\hat{F}'_0\right)+\frac{1}{\binom{n_0}{r}}\sum G\left(x^0_{i_1},...,x^0_{i_r}\right)^2+\frac{2}{\binom{n_0}{r}}\sum G\left(x^0_{i_1},...,x^0_{i_r}\right)\left(\hat{F}'_0\left(x^0_{i_1},...,x^0_{i_r}\right)-h\left(y^0_{i_1},...,y^0_{i_r}\right)\right)\\
			&\qquad+\lambda^0_{n_0}\left\lVert G\right\rVert^2_{\mathcal{H}^r_0}+2\lambda^0_{n_0}\left\langle G,\hat{F}'_0\right\rangle_{\mathcal{H}^r_0}\\
			&\geq\hat{\mathcal{E}}_\text{reg}\left(\hat{F}'_0\right)-\frac{2}{\binom{n_0}{r}}\sum G\left(x^0_{i_1},...,x^0_{i_r}\right)\left(h\left(y^0_{i_1},...,y^0_{i_r}\right)-\hat{F}'_0\left(x^0_{i_1},...,x^0_{i_r}\right)\right)+2\lambda^0_{n_0}\left\langle G,\hat{F}'_0\right\rangle_{\mathcal{H}^r_0}\\
			&=\hat{\mathcal{E}}_\text{reg}\left(\hat{F}'_0\right)-2\lambda^0_{n_0}\sum G\left(x^0_{i_1},...,x^0_{i_r}\right)c_{i_1,...,i_r}+2\lambda^0_{n_0}\sum^{n_0}_{i_1,...,i_r=1}G\left(x^0_{i_1},...,x^0_{i_r}\right)c_{i_1,...,i_r}\\
			&\qquad\qquad\text{by the reproducing property and the definition of }c_{i_1,...,i_r}\\
			&=\hat{\mathcal{E}}_\text{reg}\left(\hat{F}'_0\right)
		\end{alignat*}
		Hence, \(\hat{F}'_0\) minimises \(\hat{\mathcal{E}}_\text{reg}\) in \(\mathcal{H}_0^r\), and so \(\hat{F}'_0=\hat{F}_0\) as required. 
	\end{proof}
	\begin{customproof}{Theorem 5.2}
		Suppose \(k_0^r\) is a bounded and universal kernel and that \(\lambda^0_{n_0}\) decays at a slower rate than \(\mathcal{O}(n_0^{-1/2})\). Then as \(n_0\rightarrow\infty\),
		\[\mathbb{E}\left[\left(\hat{F}_0\left(X_1,...,X_r\right)-F_0\left(X_1,...,X_r\right)\right)^2\right]\stackrel{p}{\rightarrow}0.\]
	\end{customproof}
	\begin{proof}
		Define
		\[F_{0,\lambda^0_{n_0}}=\argmin_{F\in\mathcal{H}^r_0}\left\{\mathbb{E}\left[\left(F\left(X_1,...,X_r\right)-F_0\left(X_1,...,X_r\right)\right)^2\right]+\lambda^0_{n_0}\left\lVert F\right\rVert^2_{\mathcal{H}^r_0}\right\}.\]
		By the bias-variance decomposition, this also minimises
		\[\mathcal{E}_{\lambda^0_{n_0}}(F)=\mathbb{E}\left[\left(F\left(X_1,...,X_r\right)-h\left(Y_1,...,Y_r\right)\right)^2\right]+\lambda^0_{n_0}\left\lVert F\right\rVert^2_{\mathcal{H}^r_0}.\]
		Denote the Hilbert space of \(P^r_X\)-square-integrable \(\mathcal{X}^r\rightarrow\mathbb{R}\) functions by \(L^2(\mathcal{X}^r,P^r_X)\), and define the inclusion operator
		\[\iota:\mathcal{H}^r_0\rightarrow L^2(\mathcal{X}^r,P_X^r).\]
		Then we see that
		\begin{alignat*}{2}
			&&F_{0,\lambda^0_{n_0}}&=\argmin_{F\in\mathcal{H}^r_0}\left\{\left\lVert\iota(F)-F_0\right\rVert^2_2+\lambda^0_{n_0}\left\lVert F\right\rVert^2_{\mathcal{H}^r_0}\right\}\\
			\implies\qquad&&0&=\iota^*(\iota(F_{0,\lambda^0_{n_0}})-F_0)+\lambda^0_{n_0}F_{0,\lambda^0_{n_0}}\\
			\implies\qquad&&F_{0,\lambda^0_{n_0}}&=\left(\iota^*\circ\iota+\lambda^0_{n_0}I\right)^{-1}\iota^*F_0
		\end{alignat*}
		Now, for any \(\mathbf{x}^0=(x^0_1,...,x^0_{n_0})^T\in\mathcal{X}^{n_0}\), define the sampling operator
		\[S_{\mathbf{x}^0}:\mathcal{H}^r_0\rightarrow\mathbb{R}^{\binom{n_0}{r}},\qquad \left(S_{\mathbf{x}^0}(F)\right)_{i_1,...,i_r}=\frac{1}{\binom{n_0}{r}}F\left(x^0_{i_1},...,x^0_{i_r}\right),\{i_1,...,i_r\}\subset\{1,...,n_0\},\]
		with adjoint
		\[S^*_{\mathbf{x}^0}\left(\mathbf{h}\right)=\frac{1}{\binom{n_0}{r}}\sum k_0\left(x^0_{i_1},\cdot\right)...k_0\left(x^0_{i_r},\cdot\right)h_{i_1,...,i_r},\qquad\mathbf{h}\in\mathbb{R}^{\binom{n_0}{r}};\]
		indeed, for any \(F\in\mathcal{H}^r_0\) and \(\mathbf{h}\in\mathbb{R}^{\binom{n_0}{r}}\), 
		\begin{alignat*}{2}
			\left\langle S_{\mathbf{x}^0}F,\mathbf{h}\right\rangle_{\mathbb{R}^{\binom{n_0}{r}}}&=\frac{1}{\binom{n_0}{r}}\sum F\left(x^0_{i_1},...,x^0_{i_r}\right)h_{i_1,...,i_r}\\
			&=\frac{1}{\binom{n_0}{r}}\sum\left\langle F,k_0\left(x^0_{i_1},\cdot\right)...k_0\left(x^0_{i_r},\cdot\right)\right\rangle_{\mathcal{H}_0^r}h_{i_1,...,i_r}\\
			&=\left\langle F,\frac{1}{\binom{n_0}{r}}\sum k_0\left(x^0_{i_1},\cdot\right)...k_0\left(x^0_{i_r},\cdot\right)h_{i_1,...,i_r}\right\rangle_{\mathcal{H}^r_0}.
		\end{alignat*}
		For \(\mathbf{y}^0\in\mathcal{Y}^{n_0}\), write
		\[h\left(\mathbf{y}^0\right)\in\mathbb{R}^{\binom{n_0}{r}},\qquad h\left(\mathbf{y}^0\right)_{i_1,...,i_r}=h\left(y^0_{i_1},...,y^0_{i_r}\right),\{i_1,...,i_r\}\subset\{1,...,n_0\}.\]
		Then we see that
		\begin{alignat*}{3}
			&&\hat{F}_0&=\argmin_{F\in\mathcal{H}^r_0}\left\{\binom{n_0}{r}\left\lVert S_{\mathbf{x}^0}(F)-\frac{1}{\binom{n_0}{r}}h\left(\mathbf{y}^0\right)\right\rVert^2+\lambda^0_{n_0}\left\lVert F\right\rVert^2_{\mathcal{H}^r_0}\right\}\\
			\implies\qquad&&0&=\binom{n_0}{r}S_{\mathbf{x}^0}^*\left(S_{\mathbf{x}^0}\left(\hat{F}_0\right)-\frac{1}{\binom{n_0}{r}}h\left(\mathbf{y}^0\right)\right)+\lambda^0_{n_0}\hat{F}_0\\
			\implies\qquad&&\hat{F}_0&=\left(\binom{n_0}{r}S^*_{\mathbf{x}^0}\circ S_{\mathbf{x}^0}+\lambda^0_{n_0}I\right)^{-1}S^*_{\mathbf{x}^0}h\left(\mathbf{y}^0\right).
		\end{alignat*}
		We consider the following decomposition:
		\begin{alignat*}{2}
			\mathbb{E}\left[\left(\hat{F}_0\left(X_1,...,X_r\right)-F_0\left(X_1,...,X_r\right)\right)^2\right]=\left\lVert\iota\hat{F}_0-F_0\right\rVert^2_2&\leq2\left\lVert\iota\hat{F}_0-\iota F_{0,\lambda^0_{n_0}}\right\rVert_2^2\tag{a}\\
			&+2\left\lVert\iota F_{0,\lambda^0_{n_0}}-F_0\right\rVert^2_2.\tag{b}
		\end{alignat*}
		We are done if we show that the terms (a) and (b) separately converge to 0 (in probability, for (a)).
		\begin{enumerate}[(a)]
			\item See that
			\begin{alignat*}{2}
				\hat{F}_0-F_{0,\lambda^0_{n_0}}&=\left(\binom{n_0}{r}S^*_{\mathbf{x}^0}\circ S_{\mathbf{x}^0}+\lambda^0_{n_0}I\right)^{-1}S^*_{\mathbf{x}^0}h\left(\mathbf{y}^0\right)-F_{0,\lambda^0_{n_0}}\\
				&=\left(\binom{n_0}{r}S^*_{\mathbf{x}^0}\circ S_{\mathbf{x}^0}+\lambda^0_{n_0}I\right)^{-1}\left(S^*_{\mathbf{x}^0}h\left(\mathbf{y}^0\right)-\binom{n_0}{r}S^*_{\mathbf{x}^0}\circ S_{\mathbf{x}^0}F_{0,\lambda^0_{n_0}}+\iota^*\left(\iota F_{0,\lambda^0_{n_0}}-F_0\right)\right).
			\end{alignat*}
			By spectral theorem, 
			\[\left\lVert\hat{F}_0-F_{0,\lambda^0_{n_0}}\right\rVert_\mathcal{H}\leq\frac{1}{\lambda^0_{n_0}}\left\lVert S^*_{\mathbf{x}^0}h\left(\mathbf{y}^0\right)-\binom{n_0}{r}S^*_{\mathbf{x}^0}\circ S_{\mathbf{x}^0}F_{0,\lambda^0_{n_0}}+\iota^*\left(\iota F_{0,\lambda^0_{n_0}}-F_0\right)\right\rVert_\mathcal{H}.\]
			Using this inequality and Chebyshev's inequality, for any \(\epsilon>0\),
			\begin{alignat*}{2}
				P&\left(\left\lVert\hat{F}_0-F_{0,\lambda^0_{n_0}}\right\rVert_\mathcal{H}\geq\epsilon\right)\leq P\left(\frac{1}{\lambda^0_{n_0}}\left\lVert S^*_{\mathbf{x}^0}h\left(\mathbf{y}^0\right)-\binom{n_0}{r}S^*_{\mathbf{x}^0}\circ S_{\mathbf{x}^0}F_{0,\lambda^0_{n_0}}-\iota^*\left(F_0-\iota F_{0,\lambda^0_{n_0}}\right)\right\rVert_\mathcal{H}\geq\epsilon\right)\\
				&\leq\frac{1}{(\lambda^0_{n_0})^2\epsilon^2}\mathbb{E}\left[\left\lVert S^*_{\mathbf{x}^0}h\left(\mathbf{y}^0\right)-\binom{n_0}{r}S^*_{\mathbf{x}^0}\circ S_{\mathbf{x}^0}F_{0,\lambda^0_{n_0}}-\iota^*\left(F_0-\iota F_{0,\lambda^0_{n_0}}\right)\right\rVert_\mathcal{H}^2\right]\\
				&\leq\frac{1}{(\lambda^0_{n_0})^2\epsilon^2\binom{n_0}{r}}\mathbb{E}\left[\left\lVert k_0\left(x^0_{i_1},\cdot\right)...k_0\left(x^0_{i_r},\cdot\right)\left(h\left(y^0_{i_1},...,y^0_{i_r}\right)-F_{0,\lambda^0_{n_0}}\left(x^0_{i_1},...,x^0_{i_r}\right)\right)\right\rVert_\mathcal{H}^2\right]\\
				&\rightarrow0
			\end{alignat*}
			as \(n\rightarrow\infty\), since the kernel is bounded. 
			\item Take an arbitrary \(\epsilon>0\). By the denseness of \(\mathcal{H}^r_0\) in \(L^2(\mathcal{X}^r,P^r_X)\), there exists some \(F_\epsilon\in\mathcal{H}^r_0\) with
			\[\left\lVert\iota F_\epsilon-F_0\right\rVert^2_2=\mathcal{E}(F_\epsilon)-\mathcal{E}(F_0)\leq\frac{\epsilon}{2}.\]
			Then
			\begin{alignat*}{2}
				\left\lVert\iota F_{0,\lambda^0_{n_0}}-F_0\right\rVert^2_2&=\mathcal{E}\left(F_{0,\lambda^0_{n_0}}\right)-\mathcal{E}\left(F_0\right)\\
				&\leq\mathcal{E}_{\lambda^0_{n_0}}\left(F_{0,\lambda^0_{n_0}}\right)-\mathcal{E}\left(F_0\right)\\
				&=\mathcal{E}_{\lambda^0_{n_0}}\left(F_{0,\lambda^0_{n_0}}\right)-\mathcal{E}_{\lambda^0_{n_0}}(F_\epsilon)+\mathcal{E}_{\lambda^0_{n_0}}(F_\epsilon)-\mathcal{E}(F_\epsilon)+\mathcal{E}(F_\epsilon)-\mathcal{E}\left(F_0\right)\\
				&\leq\lambda^0_{n_0}\left\lVert F_\epsilon\right\rVert^2_{\mathcal{H}^r_0}+\frac{\epsilon}{2}.
			\end{alignat*}
			Now let \(n\) be large enough for
			\[\lambda^0_{n_0}\left\lVert F_\epsilon\right\rVert^2_{\mathcal{H}^r_0}\leq\frac{\epsilon}{2}\]
			to hold.
		\end{enumerate}
	\end{proof}
\end{document}